\documentclass[]{fairmeta}

\usepackage{amssymb}
\usepackage{colortbl}
\usepackage{adjustbox}
\usepackage{comment}
\usepackage{makecell}
\usepackage{xspace}
\usepackage[toc,page,header]{appendix}
\usepackage{minitoc}
\usepackage{url}
\usepackage{tikz}

\definecolor{metaBlue}{RGB}{24,119,242}
\definecolor{metaBlueLight}{RGB}{220,230,242}
\definecolor{darkblue}{RGB}{0,0,139}
\definecolor{americanrose}{rgb}{1.0, 0.01, 0.24}

\newcommand{\finding}[2]{
    \vspace{-0.1cm}
    \begin{tcolorbox}[
        colback=metabg,     
        arc=5pt,                    
        boxsep=5pt,                 
        left=10pt,                  
        right=10pt,                 
        top=2pt,                    
        bottom=2pt,                 
        boxrule=0.8pt,              
        drop shadow=gray!0!white,  
        enhanced jigsaw,             
        frame hidden,
    ]
    \vspace{-0.1cm}
        \paragraph{\textbf{\textit{Finding #1:}}} #2
    \vspace{-0.1cm}
    \end{tcolorbox}
    \vspace{-0.1cm}
}
 
\AtEndPreamble{
    \usepackage{cleveref}
    \crefname{section}{Sec.}{Secs.}
    \Crefname{section}{Section}{Sections}
    \crefname{table}{Tab.}{Tabs.}
    \Crefname{table}{Table}{Tables}
}

\renewcommand{\paragraph}[1]{\vspace{1.25mm}\noindent\textbf{#1}}
\newlength\savewidth

\newcolumntype{x}[1]{>{\centering\arraybackslash}p{#1pt}}
\newcolumntype{y}[1]{>{\raggedright\arraybackslash}p{#1pt}}
\newcolumntype{z}[1]{>{\raggedleft\arraybackslash}p{#1pt}}

\makeatletter
\DeclareRobustCommand\onedot{\futurelet\@let@token\@onedot}
\def\@onedot{\ifx\@let@token.\else.\null\fi\xspace}

\makeatother

\newcommand{\method}[0]{ViTok\xspace}
\usetikzlibrary{shapes.geometric, arrows.meta, positioning}
\usetikzlibrary{calc}
\tikzstyle{database} = [
    cylinder,
    shape border rotate=90,
    aspect=0.25,
    draw,
    fill=green!30!yellow!10,
    minimum height=3em,
    minimum width=3em,
    text centered,
    text width=10em 
]
\tikzstyle{process} = [
    rectangle,
    draw,
    fill=cyan!30!green!20!yellow!20,
    text centered,
    rounded corners,
    minimum height=3em,
    text width=11em 
]
\tikzstyle{decision} = [
    diamond,
    draw,
    fill=cyan!30!green!20!yellow!20,
    aspect=2,
    text centered,
    minimum width=6em,
    minimum height=3em,
    inner sep=0pt,
    text width=8em 
]
\tikzstyle{arrow} = [thick,->,>=stealth]

\title{Learnings from Scaling Visual Tokenizers for Reconstruction and Generation}

\author[1,2,\dagger]{Philippe Hansen-Estruch}
\author[2]{David Yan}
\author[2]{Ching-Yao Chung}
\author[2,4,\dagger]{Orr Zohar}
\author[2]{Jialiang Wang}
\author[2]{Tingbo Hou}
\author[2]{Tao Xu}
\author[1]{Sriram Vishwanath}
\author[2]{Peter Vajda}
\author[3]{Xinlei Chen}

\affiliation[1]{UT Austin}
\affiliation[2]{GenAI, Meta}
\affiliation[3]{FAIR, Meta}
\affiliation[4]{Stanford University}

\contribution[\dagger]{Work done at Meta}

\abstract{
Visual tokenization via auto-encoding empowers state-of-the-art image and video generative models by compressing pixels into a latent space. Although scaling Transformer-based generators has been central to recent advances, the tokenizer component itself is rarely scaled, leaving open questions about how auto-encoder design choices influence both its objective of reconstruction and downstream generative performance. Our work aims to conduct an exploration of scaling in auto-encoders to fill in this blank. To facilitate this exploration, we replace the typical convolutional backbone with an enhanced Vision Transformer architecture for Tokenization (\method). We train \method on large-scale image and video datasets far exceeding ImageNet-1K, removing data constraints on tokenizer scaling. We first study how scaling the auto-encoder bottleneck affects both reconstruction and generation -- and find that while it is highly correlated with reconstruction, its relationship with generation is more complex. We next explored the effect of separately scaling the auto-encoders' encoder and decoder on reconstruction and generation performance. 
Crucially, we find that scaling the encoder yields minimal gains for either reconstruction or generation, while scaling the decoder boosts reconstruction but the benefits for generation are mixed. Building on our exploration, we design \method as a lightweight auto-encoder that achieves competitive performance with state-of-the-art auto-encoders on ImageNet-1K and COCO reconstruction tasks (256p and 512p) while outperforming existing auto-encoders on 16-frame 128p video reconstruction for UCF-101, all with 2-5$\times$ fewer FLOPs. When integrated with Diffusion Transformers, \method demonstrates competitive performance on image generation for ImageNet-1K and sets new state-of-the-art benchmarks for class-conditional video generation on UCF-101.
}

\date{January 16, 2025}
\correspondence{Philippe Hansen-Estruch at \email{philippehansen@utexas.edu}}

\metadata[Project Page]{\url{https://vitok.github.io}}

\begin{document}
\maketitle
\section{Introduction}

\begin{figure}[t]
    \centering
    \includegraphics[width=.97\textwidth]{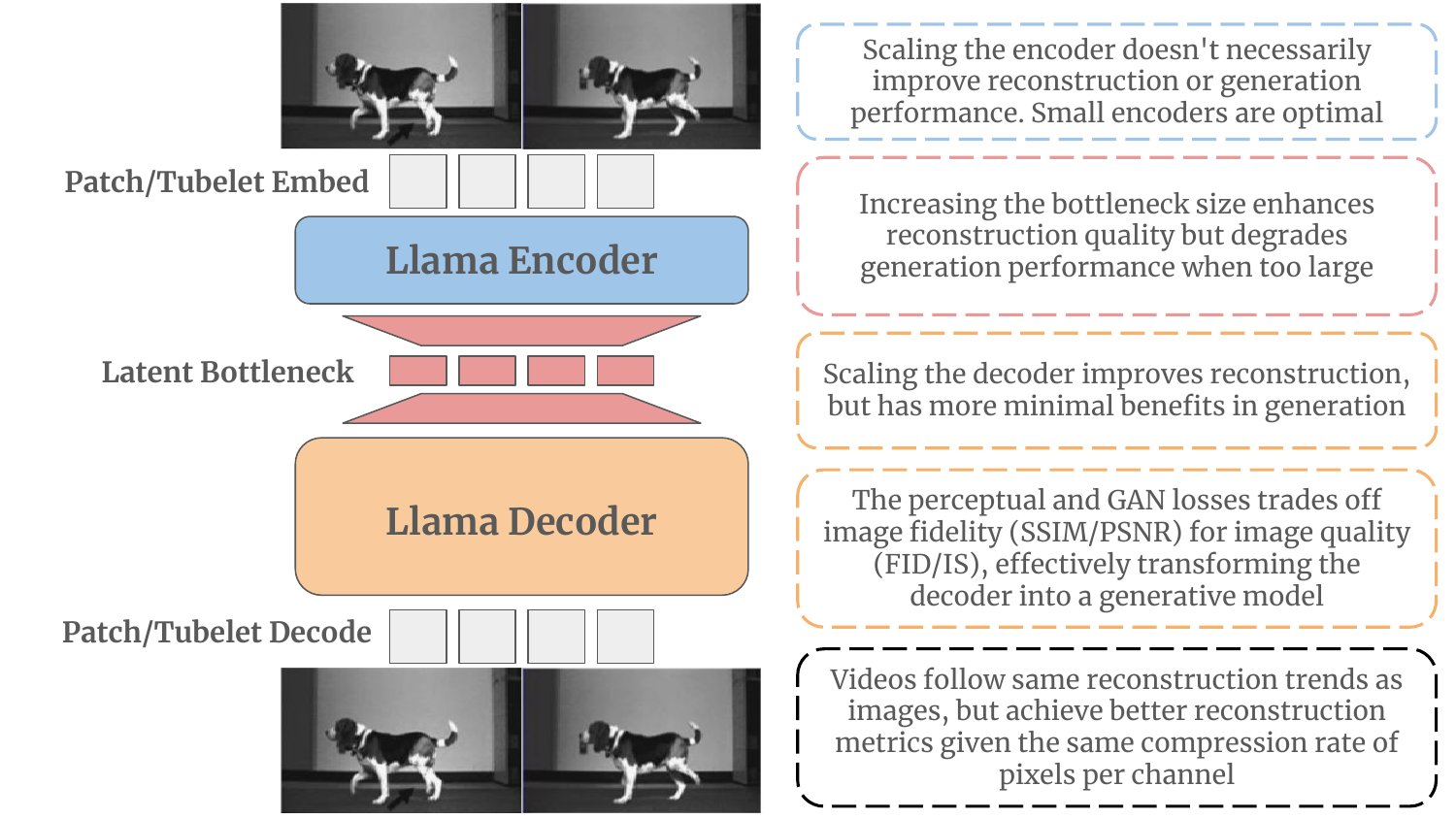}
    \caption{\textbf{Our learnings from scaling \method.} We showcase our \method architecture (left) and key findings (right) from scaling auto-encoders for image and video reconstruction and generation. We enhance traditional CNN-based auto-encoders by integrating Vision Transformers (ViTs) with an upgraded Llama architecture into an asymmetric auto-encoder framework forming \textit{Vision Transformer Tokenizer} or \method. Visual inputs are embedded as patches or tubelets, processed by a compact Llama Encoder, and bottlenecked to create a latent code. The encoded representation is then upsampled and handled by a larger Llama Decoder to reconstruct the input. Color-coded text boxes highlight the effects of scaling the encoder, adjusting the bottleneck size, and expanding the decoder. Additionally, we discuss trade-offs in loss optimization and the model's adaptability to video data. Our best performing \method variant achieves competitive performance with prior state-of-the-art tokenizers while reducing computational burden.}
    \label{fig:mainfigure}
\end{figure}

Modern methods for high-fidelity image and video generation~\citep{brooks2024video, polyak2024moviegencastmedia, genmo2024mochi, esser2024scaling} rely on two components: a visual tokenizer that encodes pixels into a lower-dimensional latent space and subsequently decodes, and a generator that models this latent representation. Although numerous works have improved the generators through scaling of Transformer-based architectures~\citep{Vaswani2017, Dosovitskiy2021}, the tokenizers themselves, predominantly based on convolutional neural networks~\citep{lecun1998gradient} (CNNs), have seldom been the focus of scaling efforts.

In this paper, we investigate whether visual tokenizers warrant the same scaling efforts as generators. To enable this, we first address two primary bottlenecks: architectural limitations and data scale. First, we replace convolutional backbones with a Transformer-based auto-encoder~\citep{Vaswani2017}, specifically adopting the Vision Transformer (ViT)~\citep{Dosovitskiy2021} architecture enhanced with Llama~\citep{touvron2023llama}, which has demonstrated effectiveness in large-scale training~\citep{gu2023mamba,sun2024autoregressive}. Our resulting auto-encoder design, which we refer to as \textit{Vision Transformer Tokenizer} or \method, combines easily with the generative pipeline in Diffusion Transformers (DiT)~\citep{Peebles2023}. Second, we train our models on large-scale, in-the-wild image datasets that significantly exceed ImageNet-1K~\citep{Deng2009} and extend our approach to videos, ensuring that our tokenizer scaling is not constrained by data limitations. Under this setup, we investigate three aspects of tokenizer scaling:

\begin{itemize}
    \item \textbf{Scaling the auto-encoding bottleneck.} Bottleneck size correlates with reconstruction metrics. However, when the bottleneck becomes large, generative performance declines due to increased channel sizes.
    
    \item \textbf{Scaling the encoder.} Although one might expect a deeper encoder to capture richer features, our findings show that scaling the encoder fails to improve outcomes and can even be detrimental. In particular, more complex latents can be harder to decode and model, reducing overall performance.
    
    \item \textbf{Scaling the decoder.} Scaling the decoder boosts reconstruction quality, but its influence on downstream generative tasks remains mixed. We hypothesize that the decoder acts in part as a generator, filling in local textures based on limited information. To confirm this, we sweep loss choices including GAN~\citep{goodfellow2014generative} and observe a trade-off between PSNR—which measures fidelity to the original image—and FID—which gauges distributional alignment but overlooks one-to-one correspondence.
\end{itemize}

Collectively, these results indicate that scaling the auto-encoder tokenizer alone is not an effective strategy for enhancing generative metrics within the current auto-encoding paradigm~\citep{Esser2021}. We also observe that similar bottleneck trends apply to video tokenizers. However, \method leverages the inherent redundancy in video data more effectively, achieving superior reconstruction metrics than for images at a fixed compression rate of pixels per channel. We summarize our findings and depict our method, \method, in Figure~\ref{fig:mainfigure}. 

Based on our sweep, we compare our best performing tokenizers to prior state-of-the-art methods. \method achieves image reconstruction and generation performance at 256p and 512p resolutions that matches or surpasses current state-of-the-art tokenizers on the ImageNet-1K \citep{Deng2009} and COCO \citep{Lin2014a} datasets, all while utilizing 2–5$\times$ fewer FLOPs. In video applications, \method surpasses current state-of-the-art methods, achieving state-of-the-art results in 16-frame 128p video reconstruction and class-conditional video generation on the UCF-101 \citep{soomro2012ucf101} dataset.

\section{Background}\label{sec:prelim}
We review background on continuous visual tokenizers and then describe \method to enable our exploration.

\subsection{Continuous Visual Tokenization}\label{sec:vae} The Variational Auto-Encoder (VAE)~\citep{Kingma2013} is a framework that takes a visual input $X \in \mathbb{R}^{T \times H \times W \times 3}$ (where $T=1$ for images and $T>1$ for videos) is processed by an encoder $f_{\theta}$, parameterized by $\theta$. This encoder performs a spatial-temporal downsampling by a factor of $q \times p \times p$, producing a latent code. The encoder outputs parameters for a multivariate Gaussian distribution—mean $z_m$ and variance $z_v$ with $c$ channel size.:
\vspace{-1mm}
$$ z \sim \mathcal{N}(z_m, z_v) = Z = f_{\theta}(X) \in \mathbb{R}^{\frac{T}{q} \times \frac{H}{p} \times \frac{W}{p} \times c}, $$

The sampled latent vector $z$ is then fed into a decoder $g_{\psi}$, with parameters $\psi$, which reconstructs the input image $\hat{X} = g_{\psi}(z)$. The primary objective of the auto-encoder is to minimize the mean squared error between the reconstructed and original images, \( \mathcal{L}_{\text{REC}}(\hat{X}, X) \). To regularize the latent distribution to a unit Gaussian prior which is necessary to recover the variational lower bound, a KL divergence regularization term is added which we refer to as \( \mathcal{L}_{\text{KL}} \). Recent advancements in VAEs used for downstream generation tasks~\citep{Esser2021, Rombach2022} incorporate additional objectives to improve the visual fidelity of the reconstructions. These include a perceptual loss based on VGG features~\citep{johnson2016perceptual} $\mathcal{L}_{\text{LPIPS}}$ and an adversarial GAN objective, $\mathcal{L}_{\text{GAN}}$~\citep{goodfellow2014generative}. The comprehensive loss function for the auto-encoder, \( \mathcal{L}_{\text{AE}}(\hat{X}, X, Z) \),  is formulated as:

\begin{equation}\label{eq:vae}
\mathcal{L}_{\text{AE}}(\hat{X}, X, Z) = \mathcal{L}_{\text{REC}}(\hat{X}, X) + \beta \mathcal{L}_{\text{KL}}(Z) + \eta \mathcal{L}_{\text{LPIPS}}(\hat{X}, X) + \lambda \mathcal{L}_{\text{GAN}}(\hat{X}, X) 
\end{equation}
where $\beta$, $\eta$, and $\lambda$ are weights that balance the contribution of each term to the overall objective. We largely utilize the same overall loss, but ablate on the impact of each term in Section~\ref{sec:tok_trade}. 
\vspace{-2mm}

\subsection{Scalable Auto-Encoding Framework}\label{sec:method}
We now develop our visual tokenizer and pinpoint bottlenecks that we explore further in Section~\ref{sec:Exploration}. The basic structure follows that of a variational auto-encoder (VAE)~\citep{Kingma2013} with an encoder-decoder architecture, but rather than relying on CNNs, we adopt a Vision Transformer (ViT)~\citep{Dosovitskiy2021} approach for better scalability. Our method builds on the ViViT framework~\citep{arnab2021vivit} to handle both images and videos. Specifically, a 3D convolution with kernel and stride size $q \times p \times p$ first tokenizes the input $X$ into a sequence $X_{\text{embed}} \in \mathbb{R}^{B \times L \times C_f}$, where $L = \frac{T}{q} \times \frac{H}{p} \times \frac{W}{p}$ and $C_f$ is the transformer’s feature dimension. A ViT encoder then processes $X_{\text{embed}}$, and a linear projection reduces the channel width to produce a compact representation $Z = f_\theta(X_{\text{embed}}) \in \mathbb{R}^{B \times L \times 2c}$.
Following the VAE formulation (Section~\ref{sec:prelim}), we recover $z \in \mathbb{R}^{B \times L \times c}$. We define
\begin{equation}\label{eq:2}
E = L \times c,
\end{equation}
which effectively controls our compression ratio by specifying the total dimensionality of the latent space. As Section~\ref{sec:Exploration} highlights, $E$ is pivotal in predicting reconstruction performance. Both $c$ and $E$ are very important for generative performance as well. Though $E$ can be influence also by the number of tokens $L$, so we can potentially keep $c$ low while increasing $L$ for increased $E$. 

For the decoder, a linear projection upsamples $z$ from $c$ to $C_g$ channels, after which a ViT decoder processes the tokens to predict $\hat{X}_{\text{embed}}$. Finally, a 3D transposed convolution recovers the original input resolution, producing $\hat{X}$. This covers the high level process of \textit{Vision Transformer Tokenizer} or \method. Figure~\ref{fig:mainfigure} illustrates this process. We denote \method configurations by specifying their encoder size, decoder size, and patch/stride parameters $(q,\,p)$. For instance, \method S-B/4x16 indicates a small encoder, a base decoder, and a patch stride of $q=4$, $p=16$. Table~\ref{tab:reference_sizes} provides details on the \method sizes.

\begin{table}[t]
\centering
\label{tab:model_sizes}
\begin{tabular}{lccccc}
\hline
\textbf{Model} & \textbf{Hidden Dimension} & \textbf{Blocks} & \textbf{Heads} & \textbf{Parameters (M)} & \textbf{GFLOPs} \\
\hline
Small (S)   & 768  & 6  & 12 & 43.3 & 11.6 \\
Base (B)    & 768  & 12 & 12 & 85.8 & 23.1 \\
Large (L)   & 1152 & 24 & 16 & 383.7 & 101.8 \\
\hline
\end{tabular}
\caption{\textbf{Model Sizes and FLOPs for \method.} We describe \method variants by specifying the encoder and decoder sizes separately, along with the tubelet sizes. For example, \method S-B/4x16 refers to a model with an encoder of size Small (S) and a decoder of size Base (B), using tubelet size $q=4$ and $p=16$. We modified the traditional Small (S) model by increasing its hidden dimension from 384 to 768 and reducing the number of blocks from 12 to 6 to increase flops and parameters slightly. Additionally, for the Large (L) model, we increased the hidden dimension to 1152 from 1024 to ensure divisibility by 3 for 3D RoPE integration.
} \label{tab:reference_sizes}
\end{table}

\subsection{Experiment Setup and Training}\label{sec:Experimental_Setup} 
We detail the training process for \method that will enable our exploration in Section~\ref{sec:Exploration}.

\paragraph{Training stages.} Due to the known instability of adversarial objectives in VAE frameworks~\citep{yu2021vector}, we stage our training of \method into two parts. Stage~1 uses only the MSE, LPIPS, and KL terms, following Equation~\ref{eq:vae} with $\beta = 1 \times 10^{-3}$, $\eta = 1.0$, and $\lambda = 0$. This setup ensures a stable auto-encoder that performs well. Stage~2 then introduces an adversarial loss~\citep{goodfellow2014generative, Esser2021}, freezing the encoder $f_\theta$ while fine-tuning only the decoder $g_{\psi}$. Here, we switch to $\beta = 1 \times 10^{-3}$, $\eta = 1.0$, and $\lambda = 1.0$ in Equation~\ref{eq:vae}. For images, this adversarial component follows standard GAN-based VAE techniques. For videos, we treat each frame independently by flattening the video into batches of frames, computing LPIPS and GAN losses on a frame-by-frame basis. This two-stage approach preserves the encoder’s stability while enabling generative refinement in the decoder.

\paragraph{Architecture, datasets, and training details.}
We employ a Vision Transformer (ViT) setup for both our encoder and decoder, drawing on several modifications from Llama~\citep{touvron2023llama}. In particular, we adopt SwiGLU~\citep{shazeer2020glu} and 3D Axial RoPE~\citep{su2024roformer} to better capture spatiotemporal relationships.

Since we aim to scale our models without being constrained by data size, we train our auto-encoders on large-scale datasets. For images, we use the Shutterstock image dataset (450M images) and ImageNet-1K~\citep{Deng2009} (1.3M images), evaluating reconstruction on the ImageNet-1K validation set and COCO-2017~\citep{Lin2014} validation set. For video training, we employ the Shutterstock video dataset (30M videos, each with over 200 frames at 24 fps), and validate on UCF-101~\citep{soomro2012ucf101} and Kinetics-700~\citep{Kay2017}.

Stage 1 training runs for 100{,}000 steps, with a batch size of 1024 for images and 256 for videos. We then finetune for Stage 2 for another 100{,}000 steps, using a reduced batch size of 256 for images and 128 for videos. We use the AdamW optimizer~\citep{Kingma2015, loshchilov2017decoupled} with $\beta_1=0.9$, $\beta_2=0.95$, a peak learning rate of $\frac{1\times10^{-4}}{256}$ (scaled by batch size $\times$ frames), a weight decay of $1\times10^{-4}$, and a cosine decay schedule~\citep{Loshchilov2016}. When a discriminator is used in Stage 2, we utilize StyleGAN~\citep{karras2019stylebasedgeneratorarchitecturegenerative} and set the discriminator learning rate to $2\times10^{-5}$, with a linear warmup of 25k steps. We use bfloat16 autocasting for all training, apply no exponential moving average (EMA) in Stage 1, and introduce EMA at 0.9999 in Stage 2.

\paragraph{Reconstruction evaluation metrics.} 
To gauge reconstruction quality, we use Fréchet Inception Distance (FID)~\citep{Heusel2017}, Inception Score (IS)~\citep{salimans2016improved}, Structural Similarity Index Measure (SSIM)~\citep{1284395}, and Peak Signal-to-Noise Ratio (PSNR). For video, we report rFID (frame-wise FID) and Fréchet Video Distance (FVD)~\citep{unterthiner2019accurategenerativemodelsvideo} over entire videos, denoted as rFID and FVD respectively. We refer to these reconstruction-specific metrics as rFID, rIS, rSSIM, and rPSNR.

\paragraph{Generation experiments and metrics.} 
To assess our tokenizers in a large-scale generative setting, we train a class-conditional DiT-L~\citep{Peebles2023} with 400M parameters for 500,000 steps and a batch size of 256, applying classifier-free guidance (CFG)~\citep{ho2022classifier} on a DDIM sampler~\citep{Song2020} over 250 steps and CFG scales of 1.5 and 3.0. We apply the same Llama upgrades to our DiT as for our tokenizers. We measure generation quality using gFID and gIS (gInception Score) computed over 50,000 samples. Since \method can directly output continuous tokens, we can feed the noised latents $z + \epsilon$ directly into DiT without patchifying and predict the noise.

\section{Bottlenecks, Scaling, and Trade-offs in Visual Tokenization}\label{sec:Exploration}
In Section~\ref{sec:prelim}, we introduced \method and outlined its training process. Here, we examine the impact of scaling three key factors—bottleneck size, encoder size, and decoder size—on both reconstruction and generation performance.  First, in Section~\ref{sec:e_anal}, we examine scaling the primary bottleneck in reconstruction: the total number of floating points $E$ (Equation~\ref{eq:2}) in the latent representation. Next, in Section~\ref{sec:gen_results}, we test how this bottleneck effects generation results. Then, in Section~\ref{sec:tok_scaling}, we analyze the impact of scaling the encoder and decoder size. Afterward, in Section~\ref{sec:tok_trade}, we analyze the decoder as an extension of the generative model and examine how the choice of objective in Equation~\ref{eq:vae} influences the trade-off in reconstruction. Finally, in Section~\ref{sec:videos}, we extend our study to video data, highlighting key similarities and differences relative to image-based auto-encoding. Unless stated otherwise, all experiments in this section use Stage 1 training from Section~\ref{sec:Experimental_Setup} to ensure stable and consistent comparisons.

\subsection{$E$ as the Main Bottleneck in Image Reconstruction}\label{sec:e_anal}

\begin{figure}[t]
    \centering
    \makebox[\textwidth][c]{%
        \includegraphics[width=1.0\textwidth]{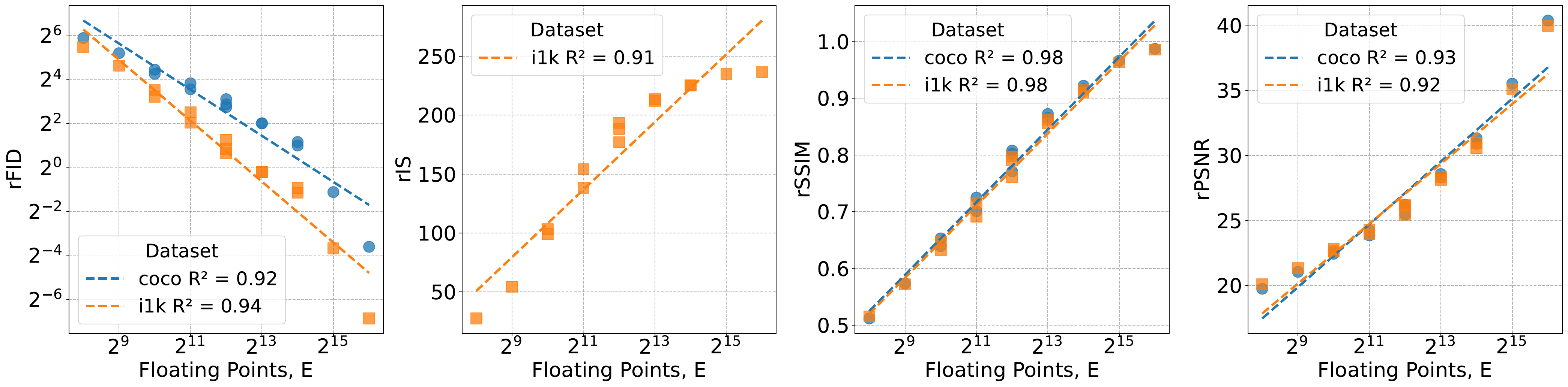}
    }
    \caption{\textbf{256p image reconstruction sweep over floating points $E$.}  We evaluate \method S-B trained with stage 1 (Section~\ref{sec:Experimental_Setup}) using combinations of patch sizes $p \in {8, 16, 32}$ and channel widths $c \in {4, 8, 16, 32, 64}$ to investigate how the total floating points $E = \frac{256^2}{p^2} \cdot c$ influences FID, IS, SSIM, and PSNR in reconstruction tasks. Our findings reveal a strong correlation between $\log(E)$ and $\log(\text{rFID})$, $\log(E)$ and $\text{rIS}$, $\log(E)$ and $\text{rSSIM}$, as well as $\log(E)$ and $\text{rPSNR}$, independent of the number of FLOPs utilized by the auto-encoder. This indicates that $E$ is the primary bottleneck for reconstruction, irrespective of the code shape or FLOPs expended. Additionally, similar trends are observed across the ImageNet-1K and COCO datasets, indicating that these patterns are consistent regardless of the dataset used. }
    \label{fig:256p_image_sweep}
\end{figure}

In prior discrete cases performance depends on the number of tokens ($L$) and the size of the discrete codebook per token~\citep{Oord2017, mentzer2023finite}. For \method, the analogous factor is $E$ (Equation~\ref{eq:2}), which proves to be the critical determinant of reconstruction performance. The bottleneck $E$ is related to the number of pixels per floating point, $\frac{T \times H \times W \times 3}{E}$, representing the degree of compression applied. 

To fully understand how $E$ functions as a bottleneck, we performed an extensive sweep through various configurations of \method investigating performance on 256p image reconstruction. For our first experiment, we look to explore all combinations of patch size $p = \{32, 16, 8\}$ and channel widths $c = \{4, 8, 16, 32, 64\}$ which gives various $E$ between $2^8$ to $2^{16}$. The patch size influences $L=\frac{H \times W}{p^2}$ and the amount of flops expended by the model due the quadratic nature of attention, while $c$ dictates the extent of the bottleneck between the encoder and the decoder. For these experiments, we fixed the encoder size to Small and the decoder to Base (Table~\ref{tab:reference_sizes}). Our findings on scaling $E$ with 256p images are summarized in Figure~\ref{fig:256p_image_sweep}. We provide more details and results in Appendix~\ref{sec:additional_experiments}.

Figure~\ref{fig:256p_image_sweep} illustrates a strong correlation between $E$ and rFID/rIS/rSSIM/rPSNR. This indicates that $E$ is a significant predictor of the quality of the reconstruction, regardless of the shape of the code. Also, the behavior between different datasets reconstruction performance is similar with rFID changing slightly due to the size of the validation set difference (50k for ImageNet-1K vs 5k for COCO). Furthermore, for the same $E$, different patch sizes ($c=\frac{E \times p^2}{H \times W}$) yield similar performance. This suggests that increasing FLOPs for a fixed $E$ does not enhance performance, establishing $E$ as the most critical bottleneck in reconstruction performance for a given encoder. Figure~\ref{fig:recon_viz} compares visualizations for different $E$ values on 256p images. As $E$ decreases, high-frequency details are lost, and when $E < 4096$, significant texture and pattern information is degraded, although the overall image structure remains intact.

One potential source of concern is the precision of $E$ could effect reconstruction performance, therefore it should be shown via bits per pixel. We train \method S-B/16 at float32 precision and compare to bfloat16 precision in Table~\ref{tab:floating_point_comparison}. There are almost no differences in performance, which shows that the
precision of $E$ does not necessarily affect the reconstruction performance. 

\newcommand{\s}{@{\hskip .5mm}}
\newcommand\recon[2]{\includegraphics[width=.15\linewidth]{recon_processed/#1_cw#2.png}}
\newcommand\truth[1]{\includegraphics[width=.15\linewidth]{recon_processed/#1_truth.png}}
\begin{figure}[t]
\centering
\begin{tabular}{c|ccccc}
\toprule
 & \multicolumn{5}{c}{Reconstruction with E Floating Points} \\
Ground Truth & 16384 & 8192 & 4096 & 2048 & 1024 \\
\midrule
\truth{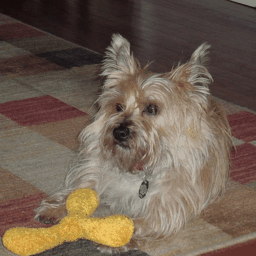} & \recon{0}{64} & \recon{0}{32} & \recon{0}{16} & \recon{0}{8} & \recon{0}{4} \\
\truth{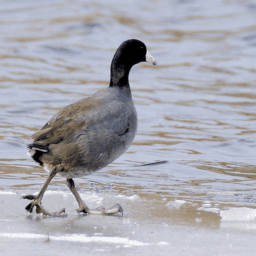} & \recon{2}{64} & \recon{2}{32} & \recon{2}{16} & \recon{2}{8} & \recon{2}{4} \\
\truth{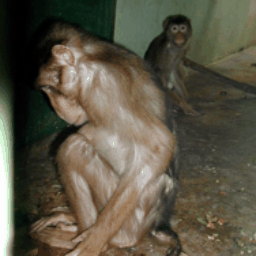} & \recon{3}{64} & \recon{3}{32} & \recon{3}{16} & \recon{3}{8} & \recon{3}{4} \\
\end{tabular}
\caption{\label{fig:recon_viz} \textbf{256p image reconstruction visualization over floating points $E$.} Example reconstructions for varying the number of floating points $E$ values on \method S-B/16, achieved by adjusting the channel size $c = {64, 32, 16, 8, 4}$ for each image across the row. As $E$ decreases, high-frequency details diminish, with small colors and fine details gradually lost. When $E < 4096$, textures merge, and significant detail loss becomes apparent.
}
\end{figure}

\begin{table}[t]
\centering
\begin{tabular}{c|c|c|c|c}
\toprule
Precision& rFID& rIS & rSSIM&rPSNR\\
\midrule
BFloat16& 1.63& 194& 0.79&26.1\\
Float32& 1.62& 194& 0.80&26.1\\
\bottomrule
\end{tabular}
\caption{\textbf{Precision comparison for $E$.} We train \method S-B/16 with full float32 precision and bfloat16 autocasting on 256p images in same fashion as Figure~\ref{fig:256p_image_sweep}. The performance is close indicating that $E$ isn't effected by changing precision.}
\label{tab:floating_point_comparison}
\end{table}

\begin{figure}[h]
\centering
\makebox[\textwidth][c]{
    \includegraphics[width=1.0\textwidth]{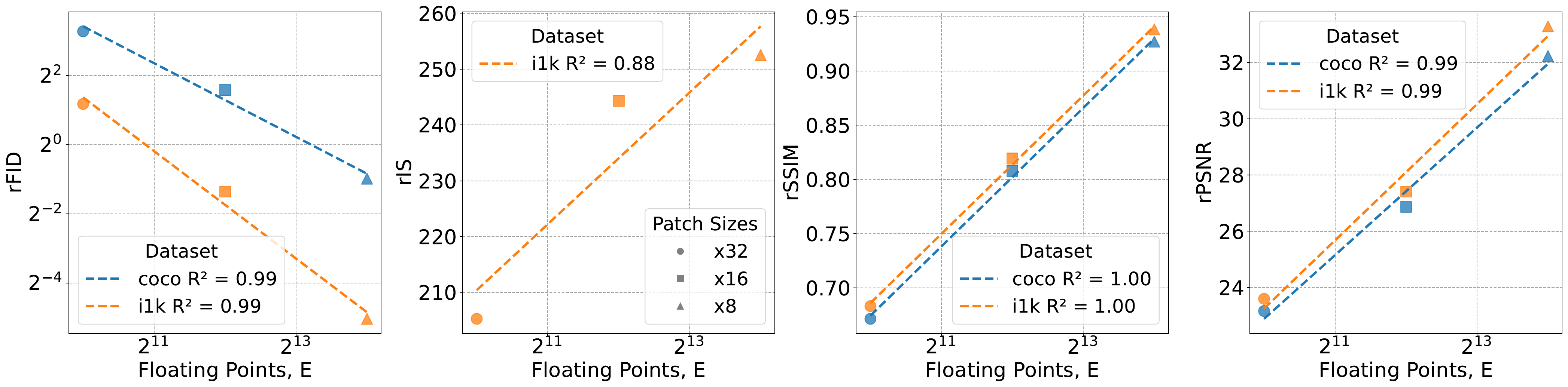}
}
\caption{\textbf{512p Image reconstruction over $E$.} We evaluate \method S-B trained with stage 1 (Section~\ref{sec:Experimental_Setup}) across all combinations of patch sizes $p \in {8, 16, 32}$ and a fixed channel width $c = 16$, analyzing how the total floating-point operations, calculated as $E = \frac{512^2}{p^2} \cdot c$, influence reconstruction metrics such as FID, IS, SSIM, and PSNR. $E$ shows trends similar to 256p results (Figure~\ref{fig:256p_image_sweep}). However, achieving comparable rPSNR/rSSIM to 256p requires $4 \times E$ for 512p reconstruction, which indicates that compression ratio of pixels to channels should be fixed to maintain performance.}
\label{fig:512p_image_sweep}
\end{figure}

\paragraph{\textbf{512p reconstruction results on total floating points $E$.}} To examine how resolution size affects $E$, we scale up the resolution from 256p to 512p. We test \method S-B/16 over $p \in {8, 16, 32}$. The results of the sweep are shown in Figure~\ref{fig:512p_image_sweep}. The results follow a trend similar to that in Figure~\ref{fig:256p_image_sweep}, with $E$ exhibiting consistent correlation relationships. While FID and IS are challenging to compare across resolutions\footnote{The InceptionV3 network used for FID and IS calculations resizes images to 299p before feature computation, leading to potential information loss during downsampling.}, achieving comparable rSSIM and rPSNR performance at 512p requires $4 \times E$ from 256p. This suggests that maintaining performance across resolutions requires preserving the same compression ratio, $\frac{H \times W \times 3}{E}$.

\finding{1}{Regardless of code shape or flops expended in auto-encoding, the total number of floating points in the latent code ($E$) is the most predictive bottleneck for visual reconstruction performance.}

\subsection{The Impact of $E$ in Image Generation}\label{sec:gen_results}
\begin{figure}[t]
\centering
\makebox[\textwidth][c]{
    \includegraphics[width=1.0\textwidth]{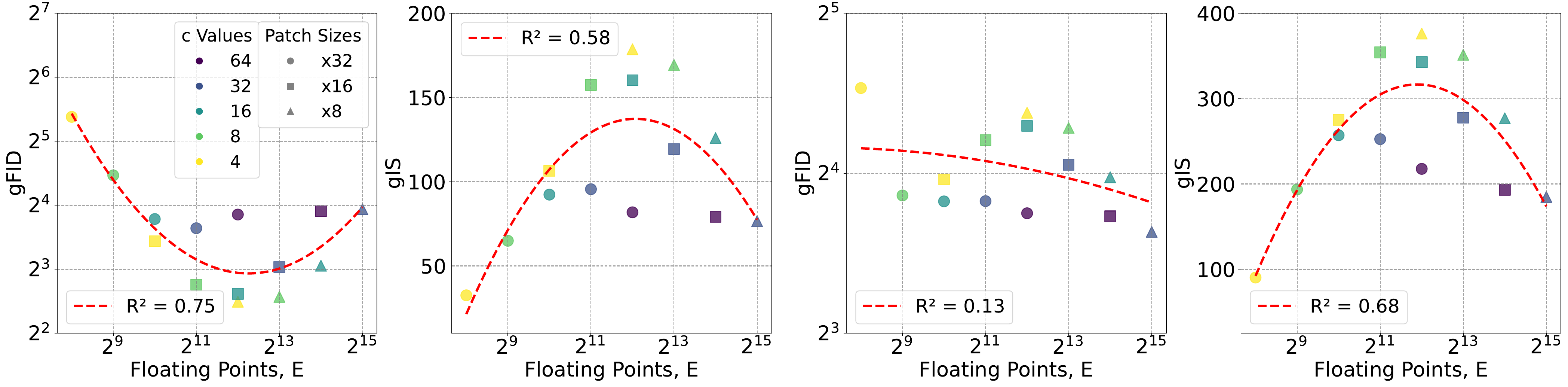}
}
\caption{\textbf{256p image generation over $E$.} We evaluate each tokenizer from Figure~\ref{fig:256p_image_sweep} on DiT following Section~\ref{sec:Experimental_Setup}. Results for CFG scales of 1.5 and 3.0 are on the left two and right two plots respectively. Our results show no strong linear correlation between $\log(E)$ and generation performance. Instead, a second-order trend reveals an optimal $E$ for each patch size $p$, indicating a complex interplay between $E$ and $c$. This highlights the necessity of optimizing both parameters to balance reconstruction quality with generative capabilities.}
\label{fig:256p_image_generation_sweep}
\end{figure}

In this section, we investigate how $E$ influences performance in generative tasks by following the training protocol from Section~\ref{sec:Experimental_Setup} and using the same set of tokenizers evaluated in Figure~\ref{fig:256p_image_sweep}. The results are in Figure~\ref{fig:256p_image_generation_sweep}.

The generative results exhibit a different trend compared to reconstruction, showing little to no linear correlation between $\log(E)$ and the generative metrics $\log(\text{gFID})$ or gIS. Figure~\ref{fig:256p_image_generation_sweep} reveals that each patch size has an optimal $E$, leading to a second-order trend. The optimal configurations are $p=16$, $c=16$, $E=4096$; $p=8$, $c=4$, $E=4096$; and $p=32$, $c=32$, $E=2048$ for their respective patch sizes. Additionally, higher CFG settings tend to minimize the differences in gFID across various $E$ values. However, for gIS, higher channel sizes ($c > 32$) and variants with poor reconstruction quality still result in poorer image quality, indicating that excessive channel sizes negatively impact performance despite CFG adjustments.

Closer analysis reveals that a low $E$ often bottlenecks the generative model, as the auto-encoder struggles with effective image reconstruction. Conversely, a high $E$, primarily driven by larger channel sizes ($c$), complicates model convergence and degrades both gFID and gIS metrics. These findings are corroborated by concurrent work that details a trade off between rFID and gFID in latent diffusion models~\citep{yao2025reconstructionvsgenerationtaming}. This highlights a critical trade-off in current latent diffusion models: $E$ and $c$ must be kept as low as possible to enhance generation performance while maintaining it high enough to ensure quality reconstructions. We provide generation visualizations for each tokenizer and trained DiT model in Appendix~\ref{sec:visualizations}.

\finding{2}{In generative tasks, scaling the number of floating points in the code ($E$) does not consistently improve generative performance. Instead, optimal results are achieved by tuning both $E$ and $c$ to balance reconstruction and generation capabilities. A low $E$ limits reconstruction quality, while high $E$ and channel size $c$ hinder the convergence and performance of the generative model.}
\subsection{Scaling Trends in Auto-Encoding}\label{sec:tok_scaling}

\begin{figure}[h]
\centering
\makebox[\textwidth][c]{
    \includegraphics[width=1.0\textwidth]{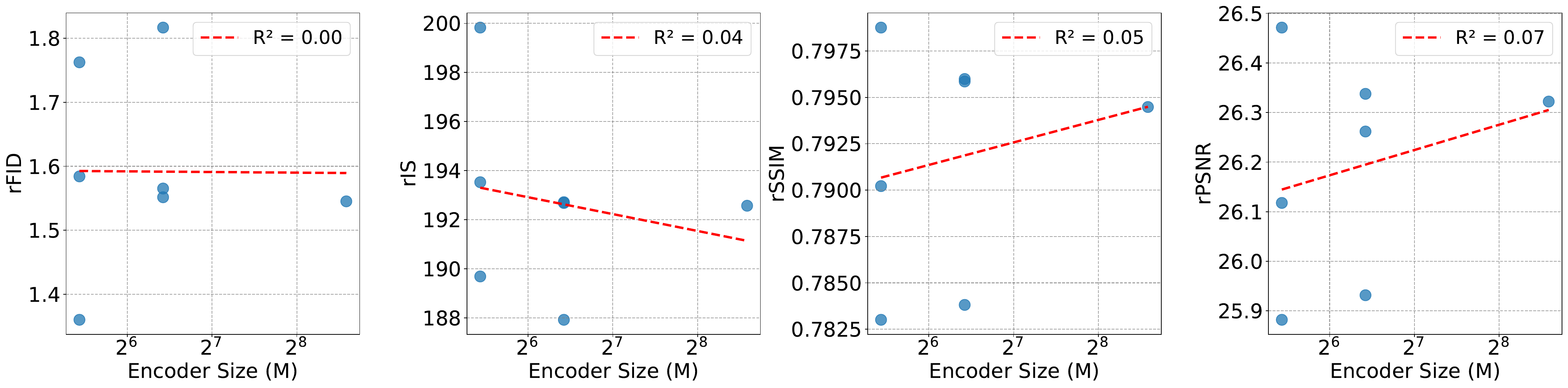}
}
\caption{\textbf{Encoder scaling on 256p image reconstruction.} We evaluate reconstruction metrics of \method trained with stage 1 (Section~\ref{sec:Experimental_Setup}) over model sizes S-S, B-S, S-B, B-B, B-L, L-L with fixed $p=16,c=16,L=256,E=4096$. There is no correlation between encoder size and reconstruction performance indicating that scaling the encoder is unhelpful in improving reconstruction capabilities. This argues that visual encoding does not require much computation.}
\label{fig:256p_image_encoder_scaling_recon}
\end{figure}

\begin{figure}[t]
\centering
\makebox[\textwidth][c]{
    \includegraphics[width=1.0\textwidth]{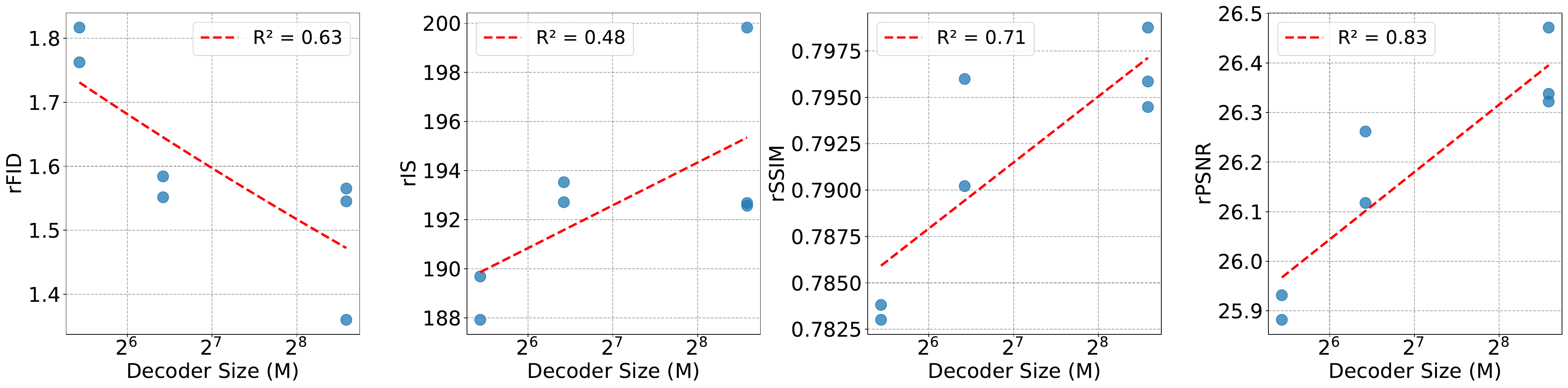}
}
\caption{\textbf{Decoder scaling on 256p image reconstruction.} Using the results from Figure~\ref{fig:256p_image_encoder_scaling_recon}, we plot various decoder sizes (S, B, L) over reconstruction performance. There is a strong correlation between decoder size and reconstruction performance, which indicates scaling the decoder improves reconstruction. Although, increasing the decoder size from Base to Large does not provide the same boost of performance as doubling $E$ to $8192$ from $4096$.}
\label{fig:256p_image_decoder_scaling_recon}
\end{figure}

We aim to explore how scaling impacts auto-encoding in both reconstruction and generation tasks using \method. To test this, we fix the parameters to $p=16, c=16, L=256, E=4096$ for \method. We then conduct a sweep over different encoder and decoder sizesS-S, B-S, S-B, B-B, S-L, B-L, L-L defined in Table~\ref{tab:reference_sizes}, following the same training protocol as described in Section~\ref{sec:Experimental_Setup}. The results are reported in Figure~\ref{fig:256p_image_encoder_scaling_recon} and~\ref{fig:256p_image_decoder_scaling_recon}.

As illustrated in Figure~\ref{fig:256p_image_encoder_scaling_recon}, the size of the encoder is not correlated with the reconstruction performance. In contrast, Figure~\ref{fig:256p_image_decoder_scaling_recon} shows that the size of the decoder is positively correlated with the reconstruction performance. However, $E$ remains the dominant factor as doubling the decoder size does not provide the same effects as doubling $E$. For example, increasing the decoder size from Base to Large drops the rFID from 1.6 to 1.3 for $E=4096$, but doubling $E$ to $8192$ brings the rFID to 0.8 (Figure~\ref{fig:256p_image_sweep}) for a decoder size Base. Overall, while scaling the decoder might be advantageous, scaling the encoder of a visual auto-encoder is unhelpful.

\begin{figure}[t]
\centering
\makebox[\textwidth][c]{
    \includegraphics[width=1.0\textwidth]{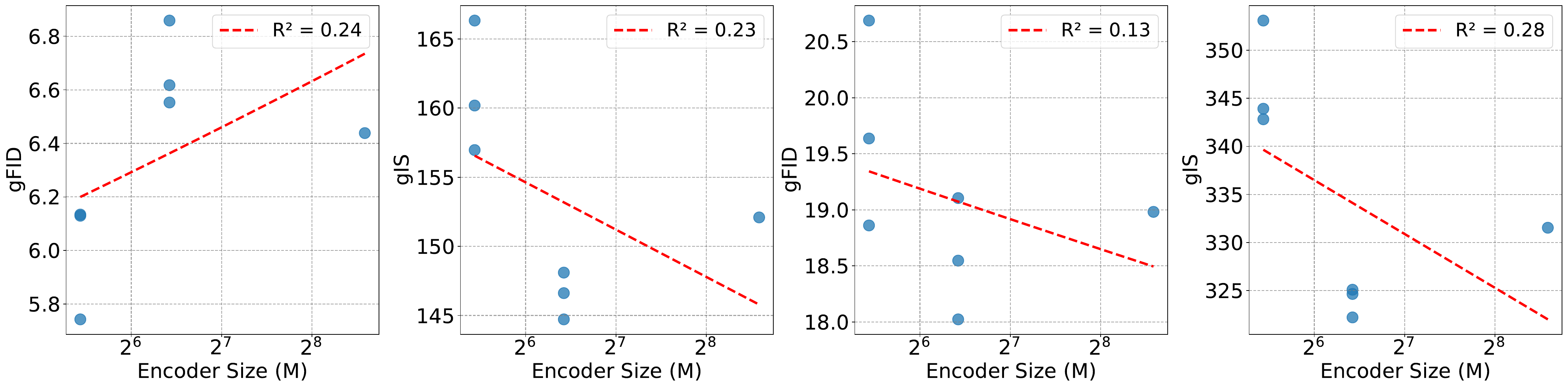}
}
\caption{\textbf{Encoder scaling on 256p image generation.} We evaluate each tokenizer from Figure~\ref{fig:256p_image_encoder_scaling_recon} on DiT following Section~\ref{sec:Experimental_Setup}. We plot encoder size over generation metric results for CFG scales of 1.5 and 3.0 on the left two and right two plots respectively. There is a weak negative correlation between encoder size and final performance indicating that scaling the encoder is harmful for generation results. This is coupled by the fact that increased encoder sizes make training slower due to increased computational overhead.}
\label{fig:256p_image_encoder_scaling_gen}
\end{figure}

\begin{figure}[t]
\centering
\makebox[\textwidth][c]{
    \includegraphics[width=1.0\textwidth]{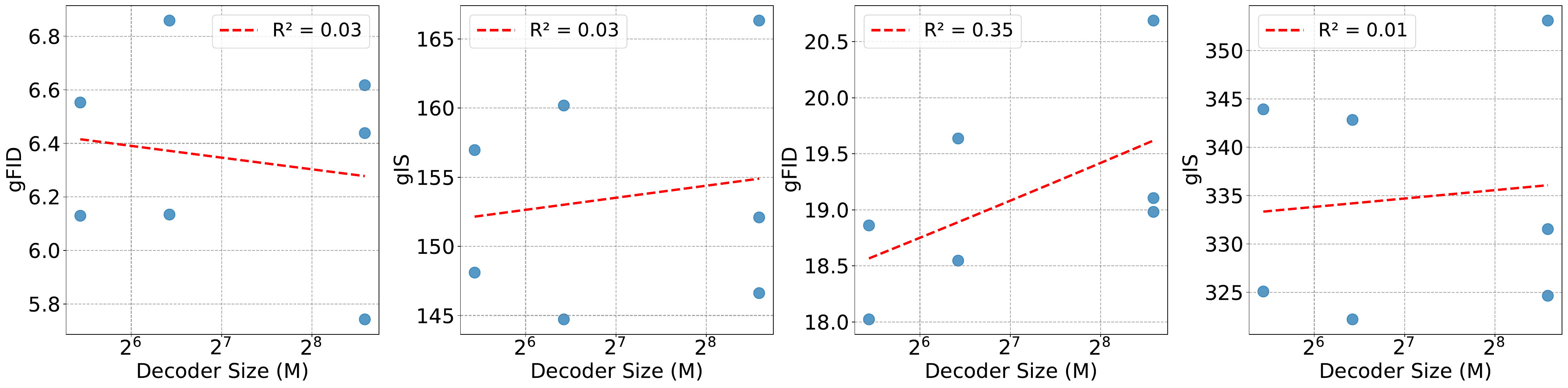}
}
\caption{\textbf{Decoder scaling on 256p image generation.} Using the results from Figure~\ref{fig:256p_image_encoder_scaling_recon}, we plot various decoder sizes (S, B, L) over generation performance. We plot decoder size over generation metric results for CFG scales of 1.5 and 3.0 on the left two and right two plots respectively. Unlike reconstruction, there is no clear correlation between decoder size and generation performance.  This indicates that scaling the decoder has minimal benefits overall for auto-encoding.}
\label{fig:256p_image_decoder_scaling_gen}
\end{figure}

Figures~\ref{fig:256p_image_encoder_scaling_gen} and~\ref{fig:256p_image_decoder_scaling_gen} explore the effects of scaling the encoder and decoder on generation performance. In Figure~\ref{fig:256p_image_encoder_scaling_gen}, a slight negative correlation is observed between encoder size and generation results. This suggests that increasing the encoder size either has little to no impact on performance or may even detrimentally affect it, all while imposing additional computational burdens. 

Similarly, Figure~\ref{fig:256p_image_decoder_scaling_gen} shows that scaling the decoder exhibits minimal correlation with generation performance, indicating that enlarging the decoder offers limited benefits. Unlike reconstruction tasks, expanding the encoder or decoder does not significantly enhance generation quality; instead, it primarily increases training and inference costs. Notably, a 129M-parameter auto-encoder performs adequately (\method S-B/16), suggesting that future scaling efforts should focus on the generation model itself rather than the auto-encoder.

\finding{3}{Scaling the encoder provides no benefits for reconstruction performance and can potentially worsen generation results.}
\vspace{1mm}
\finding{4}{While scaling the decoder can enhance reconstruction performance, it provides limited benefits for generation tasks.}

\newpage
\textbf{With the findings so far, we believe simply scaling the current auto-encoding~\citep{Esser2021} based tokenizers does not automatically lead to improved downstream generation performance. Therefore for generation, it is more cost-effective to concentrate scaling efforts on the generator itself, rather than the tokenizer.}

\begin{figure}[t]
\centering
\makebox[\textwidth][c]{
    \includegraphics[width=1.0\textwidth]{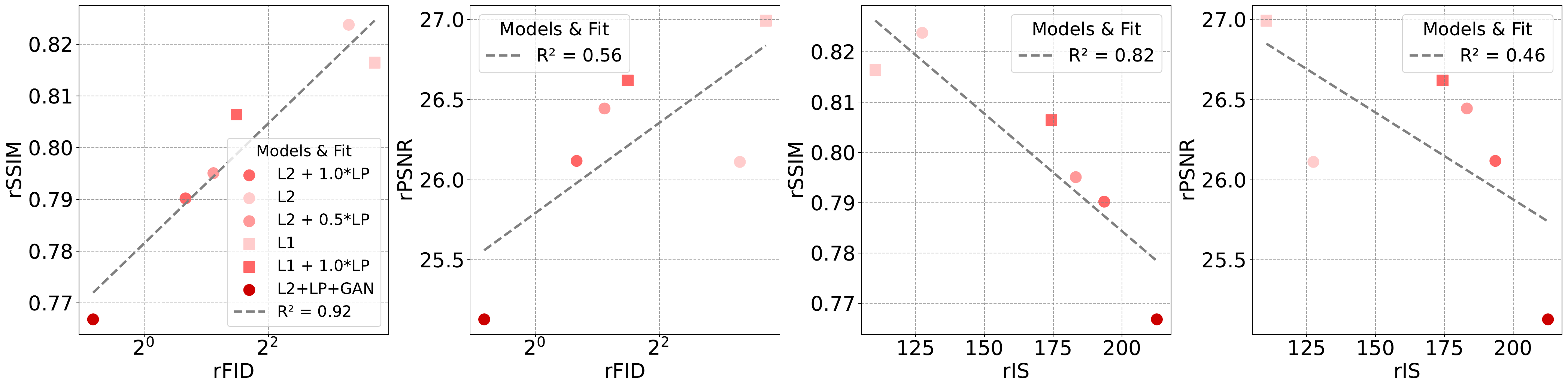}
}
\caption{\textbf{Metric trade-offs in 256p image reconstruction.} We train \method S-B/16 with stage 1 (Section~\ref{sec:Experimental_Setup}), varying the LPIPS (LP in figure) weight $\lambda \in \{0.0, 0.5, 1.0\}$ and using either L1 or L2 MSE reconstruction loss (Equation~\ref{eq:vae}). Additionally, we finetune \method S-B/16 with stage 2 and include the result as L2+LP+GAN. The results indicate that enhancing rFID/rIS scores through increased perceptual and visual losses requires a trade-off with rSSIM/rPSNR, resulting in loss of information from the original image. This indicates the decoder’s role as a generative component.
}
\label{fig:256p_image_loss_curve_tradeoff}
\end{figure}

\subsection{A Trade-Off in Decoding}\label{sec:tok_trade} As shown in Section~\ref{sec:tok_scaling}, increasing the size of the decoder improves reconstruction, suggesting that the decoder behaves more like a generative model for the input \(X\) and thus needs more computation than the encoder. To illustrate this, we compared how different losses balance traditional compression metrics (SSIM/PSNR) against generative metrics (FID/IS). SSIM/PSNR measure visual fidelity or how much of the original information is preserved, while FID/IS focus on visual quality and how closely outputs match the real dataset. This comparison shows how different choices of losses can shift the decoder’s role from strictly reconstructing to more actively generating content.

We conducted these experiments on \method by fixing \(p=16\), \(c=16\), and \(E=4096\). We then trained with stage 1 and varied the LPIPS loss weight \(\lambda \in \{0.0, 0.5, 1.0\}\) combined with the choice of L1 or L2 reconstruction loss (Equation~\ref{eq:vae}). We also include our Stage 2 results following Section~\ref{sec:Experimental_Setup} to see the effect of the generative adversarial loss.

Figure~\ref{fig:256p_image_loss_curve_tradeoff} shows a clear trade-off among these losses. Without perceptual loss, we get worse rFID/rIS scores but better rSSIM/rPSNR, indicating that a strict MSE-based approach preserves the most original information. Increasing \(\lambda\) gradually lowers SSIM/PSNR while improving FID/IS. Finally, fine-tuning the decoder with a GAN pushes these generative metrics further, achieving an rFID of 0.50 at the cost of lower SSIM/PSNR.

In addition, including the GAN also improves DiT’s downstream generation results. For instance, the Stage~1 model at \(p=16\), \(c=16\), and \(E=4096\) reaches a gFID of 5.5 and a gIS of 160 at 500k steps with CFG=1.5 (Section~\ref{sec:gen_results}). In comparison, the Stage~2 model achieves a gFID of 4.9 and a gIS of 210 at 500k steps, reflecting the same trade-offs seen in reconstruction. This underlines how strengthening the decoder’s generative capacity boosts overall performance in generation tasks, which makes the scaling benefits of decoders more complex than indicated in Section~\ref{sec:tok_scaling}.

These results demonstrate that at a fixed \(E\), aiming for higher visual quality requires sacrificing some traditional compression fidelity. This underscores that the decoder effectively acts as an extension of the generation model, creating visually pleasing results from the compressed representation. We provide more evidence of the decoder as a generative model as well as specific GAN ablations in Appendix~\ref{sec:additional_experiments}.

\finding{5}{There is a trade-off between rSSIM/rPSNR and rFID/rIS, influenced by the choice of loss weights and objectives (including perceptual and GAN losses). Consequently, the decoder can be viewed as a conditional generation model, which effectively extends the main generator.}

\subsection{Video Results}\label{sec:videos}

\begin{figure}[t] 
\centering
\includegraphics[width=\textwidth]{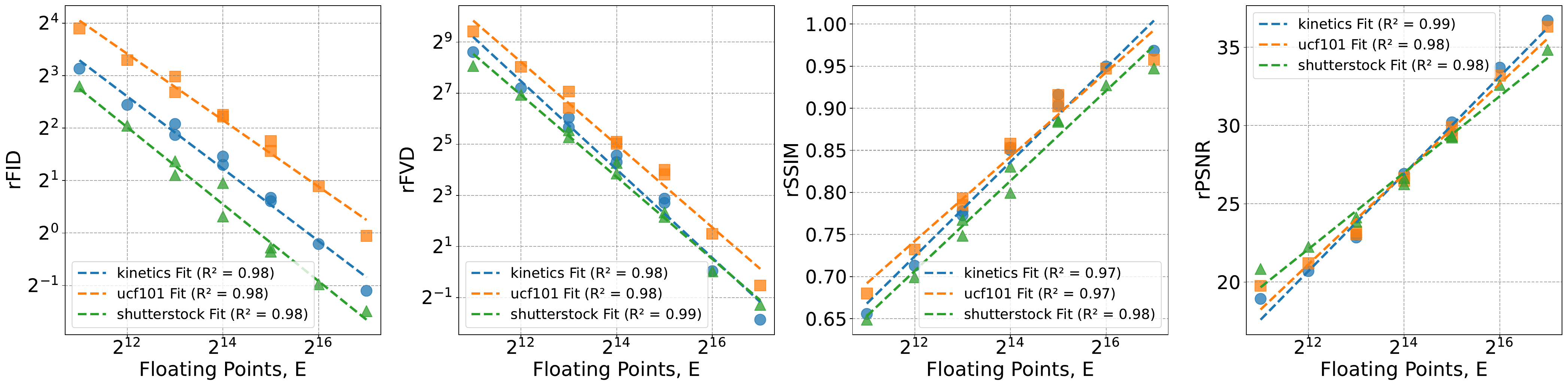}
\caption{\textbf{256p video reconstruction results over $E$.} We train \method S-B with stage 1 (Section~\ref{sec:Experimental_Setup}) on 16$\times$256$\times$256 videos at 8 fps, varying tubelet patch sizes $p \in \{8, 16, 32\}$ and temporal strides $q \in \{1, 2, 4, 8\}$ with a channel size $c=16$. Reconstruction performance is evaluated using rFID per frame, rFVD, rSSIM, and rPSNR on the Kinetics-700 validation, UCF101 training, and Shutterstock validation datasets. The results exhibit a similar trend to image reconstruction in Figure~\ref{fig:256p_image_sweep}, demonstrating a strong correlation between $E$ and reconstruction performance. Expectantly, videos are more compressible than a direct scaling from images would suggest; instead of requiring 16$\times$$E$, achieving comparable rFID, rSSIM, and rPSNR to 256p image reconstruction only necessitates 4–8$\times$$E$.}
\label{fig:E_video}
\end{figure}

\begin{figure}[t] 
\centering
\includegraphics[width=\textwidth]{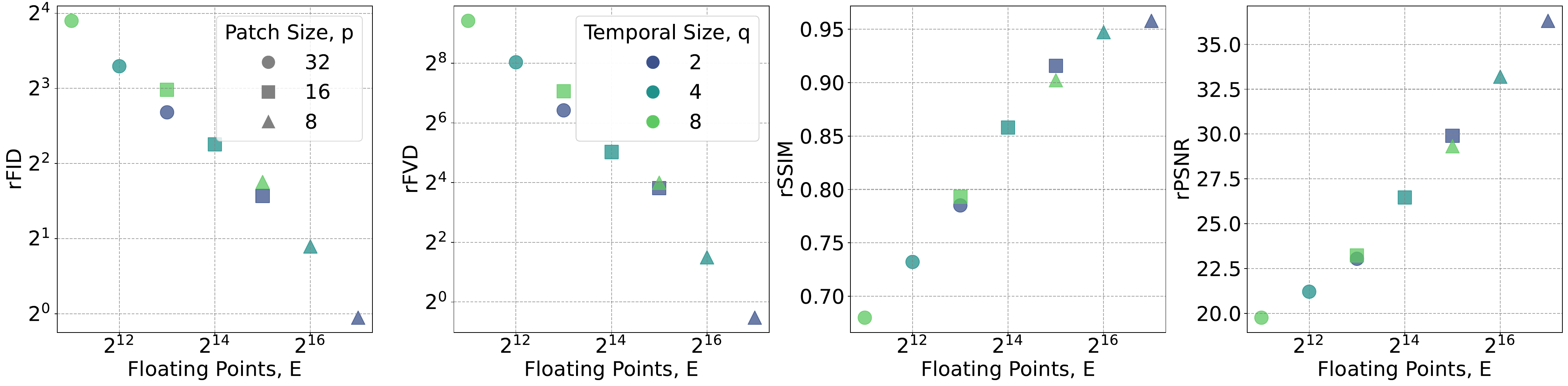}
\caption{\textbf{56p video reconstruction results detailed over $E$.} We label patch and tubelet sizes from tokenizers trained in Figure~\ref{fig:E_video}, we focus on just UCF-101 dataset due to its higher motion. For equivalent $E$, lower temporal strides are slightly more effective for better results but overall there is little benefit in trading off temporal stride for patch size in \method for videos. $E$ is still the dominating factor in predicted reconstruction performance.}
\label{fig:linear_reg_fvd}
\end{figure}

\begin{figure}[t] 
\centering
\includegraphics[width=\textwidth]{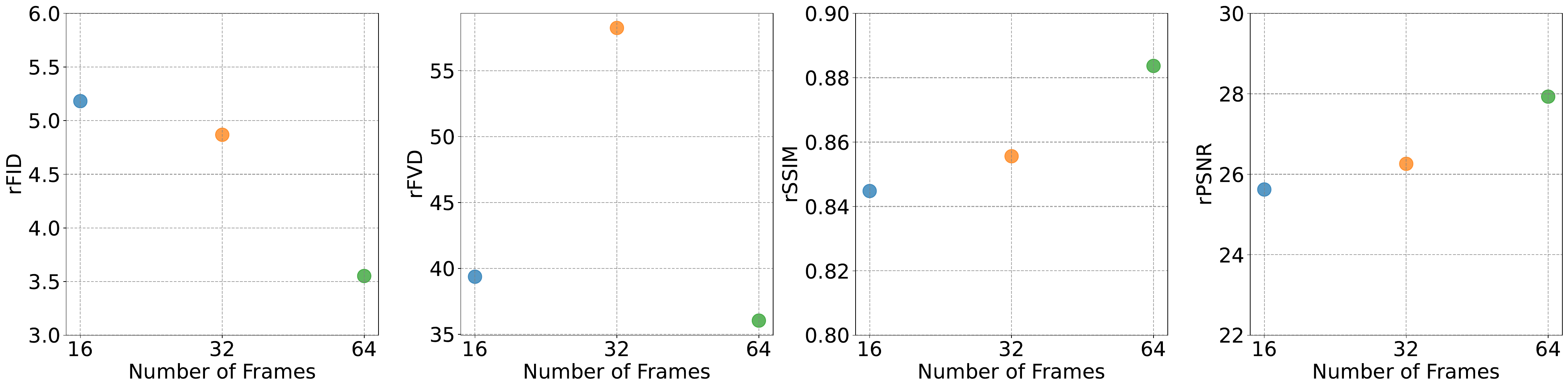}
\caption{\textbf{Multi-frame 256p video reconstruction.} We train \method S-B/4x16 with stage 1 (Section~\ref{sec:Experimental_Setup}) on 16-, 32-, and 64-frame 256p videos and evaluate reconstruction metrics on the UCF-101 dataset. The results indicate that increasing the number of frames generally improves performance, demonstrating that \method\ leverages higher redundancy in videos to achieve more efficient relative compression with same compression ratio or pixels per channel $\frac{T\times H \times W \times 3}{E}$.}
\label{fig:frame_length_ablation}
\end{figure}

We extend the application of \method to video tasks to examine the impact of $E$ on video reconstruction and to investigate redundancy in video data. To enable a direct comparison with our image results, we maintain a resolution of 256p and utilize 16-frame videos at 8 fps for both training and evaluation. Tokenizing videos can result in very large sequence lengths; for example, a tubelet size of 4×8 (with temporal stride $q=4$ and spatial stride $p=8$) for a video of dimensions 16×256×256 yields a sequence length of 4096 tokens. Therefore, based on our previous analysis of encoder and decoder sizes in Section~\ref{sec:tok_scaling}, we use a small \method S-B variant to reduce computational burden, as $E$ is likely the more critical factor in this context. 

To test how $E$ effects video we sweep over patch sizes $p \in \{8, 16, 32\}$ and temporal strides $q \in \{1, 2, 4, 8\}$ following the protocol depicted in Section~\ref{sec:Experimental_Setup}. As illustrated in Figure~\ref{fig:E_video}, the relationship between $E$ and the metrics rFVD/rFID mirrors the patterns observed in image tasks (Figure~\ref{fig:256p_image_sweep}), where $\log(E)$ strongly correlates with reconstruction metrics. Figure~\ref{fig:linear_reg_fvd} focuses on the UCF-101 dataset and demonstrates that, regardless of the selected spatial or temporal stride, $E$ remains the predominant factor influencing reconstruction performance. Consequently, adjusting spatial or temporal compression offers minimal advantages when $E$ is held constant for video reconstruction.

Comparing videos to images reveals that reaching similar rFID values requires $E \approx 16384$ to $E \approx 32768$ to achieve an rFID of 2.0, whereas for images $E=4096$ suffices. This difference, which is smaller than the naive 16$\times$ factor from frame-by-frame considerations, highlights that videos are more compressible than individual frames, and showing how \method can leverage this advantage.
\finding{6}{Videos exhibit the same reconstruction bottleneck characteristics with respect to $E$ as images do. However, auto-encoding takes advantage of the inherent compressibility of videos, enabling $E$ to scale more effectively relative to the total number of pixels than images.}

\paragraph{Scaling frame count in video reconstruction.} In our second experiment, we train \method S-B/4x16 on longer video sequences to investigate how reconstruction metrics scale with the number of frames. This analysis aims to determine whether videos become more compressible as their length increases, given that for a fixed tubelet size $E$, compression scales proportionally with the number of frames. We evaluate reconstruction performance for sequence lengths $T \in \{16, 32, 64\}$ following the training protocol outlined in Section~\ref{sec:Experimental_Setup}. As shown in Figure~\ref{fig:frame_length_ablation}, the metrics improve slightly with an increasing number of frames. This suggests that longer videos are more compressible and \method is able to take advantage of it, as the relative compression ratio (pixels per channel), calculated by $\frac{T \times H \times W \times 3}{E}$, remains constant for each $T$.

\finding{7}{Increasing the frame count for a fixed tubelet size yields improved metrics, indicating the potential for more efficient compression in longer videos.}

\section{Experimental Comparison}

\begin{figure}[t]
    \centering
    \includegraphics[width=\textwidth]{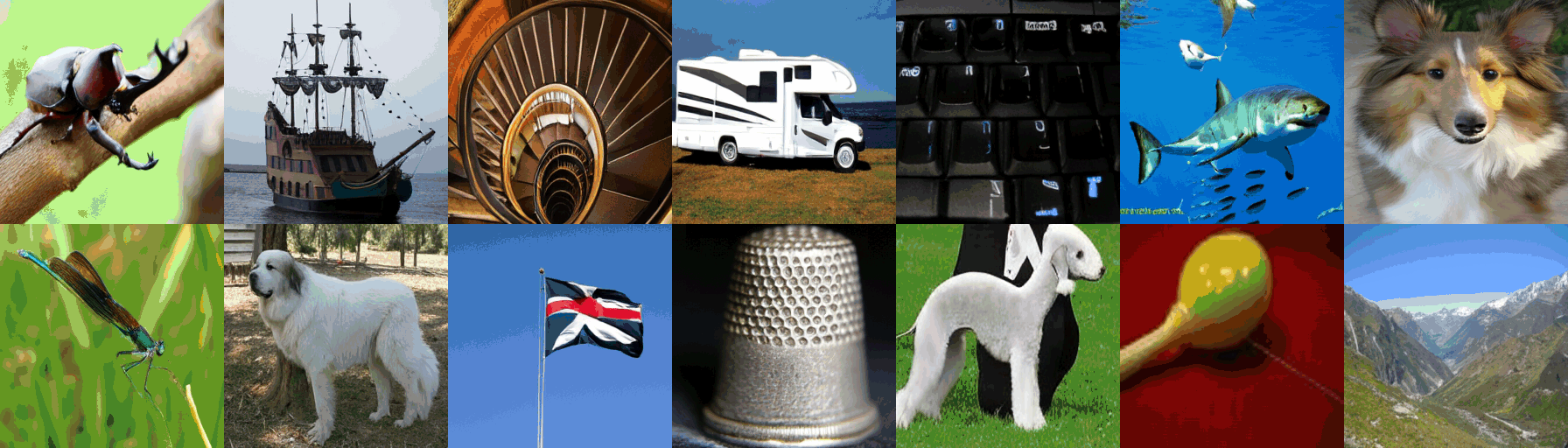}
    \caption{\textbf{256p image generation examples.} We show randomly selected 256p image generation examples from our DiT-XL trained using the \method S-B/16 variant for 4 million steps at a batch size of 256. Images were sampled with 250 steps using the DDIM sampler and a CFG weight of 4.0.}
    \label{fig:generation_visual}
\end{figure}

\begin{figure}[t]
    \centering
    \includegraphics[width=\textwidth]{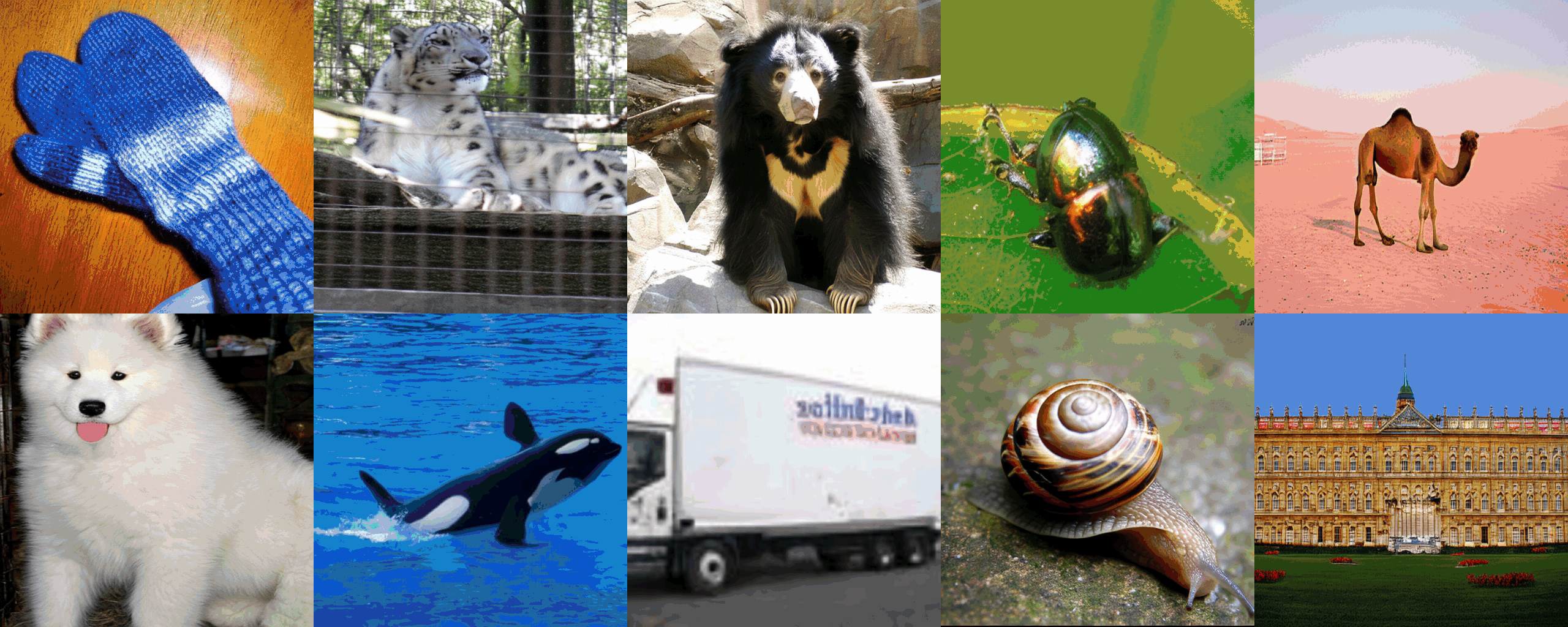}
    \caption{\textbf{512p image generation examples.} We show randomly selected 512p image generation examples from our DiT-XL trained using the \method S-B/16 variant for 4 million steps at a batch size of 256. Images were sampled with 250 steps using the DDIM sampler and a CFG weight of 4.0.}
    \label{fig:generation_visual_512}
\end{figure}

\begin{figure}[t]
    \centering
    \includegraphics[width=\textwidth]{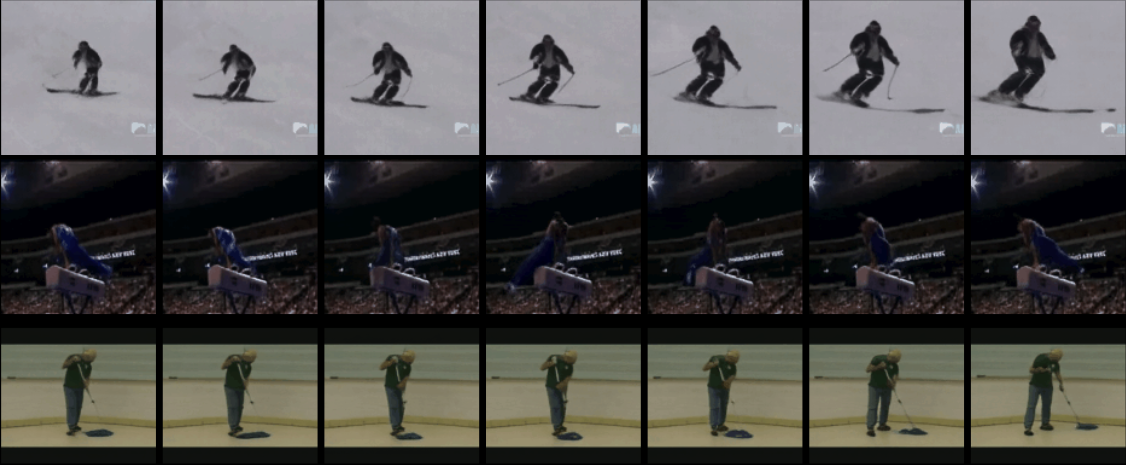}
    \caption{\textbf{128p video generation examples.} We show randomly selected 16$\times$128$\times$128 video generation examples from our DiT-L trained with \method S-B/4x8 variant. Videos are sampled with 250 steps and a CFG weight of 2.0.}
    \label{fig:vid_1024_visual}
\end{figure}

\begin{table}[t]
\centering
\begin{tabular}{ccc|ccc|ccc}
\toprule
\multirow{2}{*}{Name} & 
\multirow{2}{*}{Params (M)} & 
\multirow{2}{*}{GFLOPs} & 
\multicolumn{3}{c|}{ImageNet} & 
\multicolumn{3}{c}{COCO}\\ & & & rFID$\downarrow$ & PSNR$\uparrow$ & SSIM$\uparrow$ & rFID$\downarrow$ & PSNR$\uparrow$ & SSIM$\uparrow$ \\
\midrule
SD-VAE & 59.3 & 162.2 & 0.78 & 25.08 & 0.705 & 4.63 & 24.82 & 0.720\\
SDXL-VAE & - & -  & 0.68 & 26.04 & \textbf{0.834} & 4.07 & 25.76 & \textbf{0.845}\\
OAI & - & - & 0.81&  24.43 & 0.786 & 4.59 & 24.19 & 0.800\\
Cosmos-CI & - & - & 2.02 & \textbf{31.74} & 0.700 & 5.6 & \textbf{31.74} & 0.703\\
\midrule
\method S-B/16 & 129.0 & 34.8 & 0.50 & 24.36 & 0.747 & 3.94 & 24.45 & 0.759 \\
\method S-L/16 & 426.8 & 113.4 & \textbf{0.46} & 24.74 & 0.758 & \textbf{3.87} & 24.82 & 0.771 \\
\bottomrule
\end{tabular}
\caption{\textbf{256p image reconstruction comparison.} We assess the reconstruction performance of \method on the 256p ImageNet-1K and COCO-2017 validation sets, benchmarking them against CNN-based tokenizers with an equivalent compression ratio ($\times 16$ spatial compression). Our \method S-B/16 tokenizer achieves state-of-the-art (SOTA) rFID scores on both ImageNet-1K and COCO datasets, outperforming other CNN-based continuous tokenizers while utilizing significantly fewer FLOPs. Furthermore, \method maintains competitive performance in SSIM and PSNR metrics compared to prior methods. When scaling decoder size to Large, \method improves all its reconstruction numbers.}
\label{tab:image_recon_results}
\end{table}

\begin{table}[t]
\centering
\begin{tabular}{ccc|ccc|ccc}
\toprule
\multirow{2}{*}{Name} & 
\multirow{2}{*}{Params(M)} & 
\multirow{2}{*}{GFLOPs} & 
\multicolumn{3}{c|}{ImageNet} & 
\multicolumn{3}{c}{COCO}\\
& & & rFID$\downarrow$ & PSNR$\uparrow$ & SSIM$\uparrow$ & rFID$\downarrow$ & PSNR$\uparrow$ & SSIM$\uparrow$ \\
\midrule
SD-VAE & 59.3 & 653.8 & 0.19& - & - & - & - & - \\
\midrule
\method S-B/16 & 129.0 & 160.8 & \textbf{0.18} & 26.72 & 0.803 & \textbf{2.00} & 26.14 & 0.790\\
\bottomrule
\end{tabular}
\caption{\textbf{512p image reconstruction comparison.} We assess the reconstruction performance of our top-performing tokenizers on the 512p ImageNet-1K and COCO-2017 validation sets, benchmarking them against a CNN-based tokenizer with an equivalent compression ratio ($\times 16$ spatial compression). Our \method S-B/16 tokenizer maintains state-of-the-art (SOTA) results across all metrics, while maintaining computational significantly reducing flops.}
\label{tab:512_image_recon_results} 
\end{table}

\begin{table}[t]
\centering
\begin{tabular}{cccc|cccc}
\toprule
Method &
Params(M) & 
GFLOPs & 
\# Tokens & 
rFID$\downarrow$ & rFVD$\downarrow$ & PSNR$\uparrow$ & SSIM$\uparrow$ \\
\midrule
TATS & 32 & Unk & 2048 & - & 162 & - & -\\
MAGViT & 158 & Unk & 1280 & - & 25 & 22.0 & .701\\
MAGViTv2 & 158 & Unk & 1280 & - & 16.12 & - & - \\
LARP-L-Long & 174 & 505.3 & 1024 & - & 20 & - & -  \\
\midrule
\method S-B/4x8 & 129 & 160.8 & 1024 & 2.13 & 8.04 & 30.11 & 0.923 \\
\method S-B/8x8 & 129 & 73.2 & 512 & 2.78 & 20.05 & 28.55 & 0.898 \\
\method S-B/4x16 & 129 & 34.8 & 256 & 4.46 & 53.98 & 26.26 & 0.850 \\
\bottomrule
\end{tabular}
\caption{\textbf{128p Video Reconstruction.} We evaluate S-B/4x8, S-B/8x8, and S-B/4x16 on video reconstruction for 16$\times$128$\times$128 video on UCF-101 11k train set. \method S-B/4x8 achieves SOTA performance in rFVD and various compression statistics. \method S-B/8x8 and \method S-B/4x16 also provide competitive reconstruction numbers for the compression rate performed. \method also reduces the total FLOPs required from prior transformer based methods.} 
\label{tab:128_video_recon_results}
\end{table}

In this section, we compare our auto-encoders to prior work on image reconstruction at resolutions of 256p and 512p, as well as video reconstruction with 16 frames at 128p. We utilize the S-B/16 and S-L/16 \method variants for image tasks and the S-B/4x8, S-B/4x16, and S-B/8x8 \method variants for video tasks, as detailed in Table~\ref{tab:reference_sizes}. Training these tokenizers follows the Stage 1 and Stage 2 protocol outlined in Section~\ref{sec:Experimental_Setup}.

\subsection{Image Reconstruction and Generation}

We evaluate our models on image reconstruction and class-conditional image generation tasks using the ImageNet-1K~\citep{Deng2009} and COCO-2017 datasets at resolutions of 256p and 512p. For image reconstruction, we compare our continuous tokenizer-based models against several state-of-the-art methods, including SD-VAE 2.x~\citep{Rombach2022}, SDXL-VAE~\citep{podell2023sdxl}, Consistency Decoder~\citep{consistencydecoder}, and COSMOS~\citep{nvidia_cosmos_tokenizer}. It is important to note that discrete tokenizers present challenges for direct comparison with continuous tokenizers; therefore, our focus remains primarily on continuous tokenizers.

As shown in Table~\ref{tab:image_recon_results}, our S-B/16 variant demonstrate highly competitive performance, achieving state-of-the-art (SOTA) rFID scores on both ImageNet-1K and COCO datasets. Furthermore, our models maintain competitive metrics in rSSIM and rPSNR. When scaling up the decoder size to L, the metrics improve further showing how scaling the decoder can be helpful in \method for reconstruction. Most importantly both variants of \method reduce the required FLOPs over prior CNN methods, which highlights the efficiency of \method. For the 512p image reconstruction results presented in Table~\ref{tab:512_image_recon_results}, \method achieves SOTA reconstruction performance with a notable reduction in computational FLOPs over prior state of the art methods. In general, \method performs strongly in reconstruction benchmarks compared to prior methods in both FLOPs and performance. 

Subsequently, we assess our auto-encoders on class-conditional image generation tasks at both 256p and 512p resolutions using the ImageNet-1K dataset. We follow the DiT training protocol outlined in Section~\ref{sec:Experimental_Setup}, where we train a DiT-XL (675M parameter) model for 4 million steps paired with \method S-B/16 using 256 tokens for 256p generation and 1024 tokens for 512p generation. The results, summarized in Table~\ref{tab:image_gen_results}, indicate that \method maintains competitive performance compared to the traditional SD-VAE trained with DiT and other continuous tokenizers in image generation. In 512p generation, \method performs on par with other methods, demonstrating \method's efficacy at higher resolutions. Examples of generated images using our 256p and 512p tokenizers are illustrated in Figures~\ref{fig:generation_visual} and~\ref{fig:generation_visual_512}, respectively.

\begin{table}[t]
\centering
\begin{tabular}{ccc|cc|cc}
\toprule
\multirow{2}{*}{Tokenizer} & 
\multirow{2}{*}{Generator} & 
\multirow{2}{*}{Params (M)} & 
\multicolumn{2}{c|}{\textbf{256p Generation}} & 
\multicolumn{2}{c}{\textbf{512p Generation}} \\
\cmidrule(lr){4-5} \cmidrule(lr){6-7}
 & & & gFID$\downarrow$ & gIS $\uparrow$ & gFID$\downarrow$ & gIS $\uparrow$ \\
\midrule
SD-VAE & LDM-4 & 400 & 3.60 & 247.7 & - & - \\
SD-VAE & DiT-XL/2 & 675 & 2.27 & 278.24 & 3.04 & 240.82 \\
Taming-VQGAN & Taming-Transformer & 1400 & 15.78 & - & - & - \\
TiTok-B & MaskGIT-ViT & 177 & 2.48 & - & 2.49 & - \\
\midrule
\method S-B/16 & DiT-XL & 675 & 2.45 & 284.39 & 3.41 & 251.46 \\
\bottomrule
\end{tabular}
\caption{\textbf{Class Conditional Image Generation Results.} We evaluate our tokenizers on class-conditional generation at resolutions of 256p and 512p on the ImageNet-1K dataset compared to prior methods. \method performance is competitive with prior continuous diffusion geneation methods like SD-VAE + DiT for both 256p and 512p generation.}
\label{tab:image_gen_results}
\end{table}

\subsection{Video Reconstruction and Generation}

 For our video comparison, our reconstruction metrics are computed on the UCF-101 training set and compared against state-of-the-art methods including TATS~\citep{ge2022longvideogenerationtimeagnostic}, LARP~\citep{wang2024larp}, and MAGViTv1/v2~\citep{yu2023language,yu2023magvit}. The results are presented in Table~\ref{tab:128_video_recon_results}. Our tokenizers demonstrate very competitive performance relative to prior work. Specifically, S-B/4x8 (\(1024\) tokens) achieves state-of-the-art (SOTA) rFVD results compared to other CNN-based continuous tokenizers with the same total compression ratio. When applying further compression, the rFVD metrics show a slight degradation; however, they remain highly competitive with existing methods. Notably, our S-B/8x8 (\(512\) tokens) variant matches the performance of LARP~\citep{wang2024larp}, which operates with \(1024\) tokens. Additionally, our approach significantly reduces FLOPs compared to Transformer-based prior method LARP, underscoring the efficiency and versatility of \method.

We further evaluate our models on class-conditional video generation using the UCF-101 dataset. We train a DiT-L model across all compression variants for 500K steps on the UCF-101 training set, computing gFID and gFVD metrics with a batch size of 256 and a learning rate of \(1 \times 10^{-4}\). The results are summarized in Table~\ref{tab:video_gen_results}. \method achieves SOTA gFVD scores at \(1024\) tokens and maintains highly competitive gFVD scores at \(512\) tokens ($\times 8$ by $\times 8$ compression), representing the most efficient level of token compression for any tokenizer so far. At \(256\) tokens, \method's performance experiences a further decline but remains competitive within the field. Example video generations using our \(1024\)-token configuration are illustrated in Figure~\ref{fig:vid_1024_visual}.

\begin{table}[t]
\centering
\begin{tabular}{cccc|cc}
\toprule
Tokenizer & 
Generator & 
\# Tokens & 
Params & 
gFID$\downarrow$ & 
gFVD$\downarrow$  \\
\midrule
TATS & AR-Transformer & 2048 & 321M & - & 332\\
MAGViT & MASKGiT & 1280 & 675M & - & 76\\
MAGViTv2 & MASKGiT & 1280 & 177M & - & 58\\
W.A.L.T & DiT & 1280 & 177M & - & 46 \\
LARP-L-Long & AR-Transformer &1024 & 177M & - & 57\\
\midrule
\method S-B/4x8 & DiT &1024 & 400M & 6.67 & 27.44\\
\method S-B/8x8 & DiT &512 & 400M & 8.37 & 52.71\\
\method S-B/4x16 & DiT & 256 & 400M & 10.52 & 92.46\\
\bottomrule
\end{tabular}
\caption{\textbf{128p class conditional video generation.} We evaluate our tokenizers on class-conditional generation 16$\times$128$\times$128 on the UCF-101 dataset compared to prior methods. \method S-B/4x8 achieves SOTA performance when used with a comparable compression ratio with prior methods, though even our more aggressive tokenizer variant \method S-B/8x8 achieves SOTA results compared to prior methods. }
\label{tab:video_gen_results}
\vspace{-5mm}
\end{table}

\section{Related Work\label{sec:related}}

\paragraph{Image tokenization.} High-resolution images have been compressed using deep auto-encoders~\citep{Hinton2012, Vincent2008}, a process that involves encoding an image into a lower-dimensional latent representation, which is then decoded to reconstruct the original image. Variational auto-encoders (VAEs)~\citep{Kingma2013} extend this concept by incorporating a probabilistic meaning to the encoding. VQVAEs~\citep{Oord2017} introduce a vector quantization (VQ) step in the bottleneck of the auto-encoder, which discretizes the latent space. Further enhancing the visual fidelity of reconstructions, VQGAN~\citep{Esser2021} integrates adversarial training into the objective of VQVAE. RQ-VAE~\citep{lee2022autoregressiveimagegenerationusing} modifies VQVAE to learn stacked discrete 1D tokens. Finally, FSQ~\citep{mentzer2023finite} simplifies the training process for image discrete tokenization to avoid additional auxiliary losses.

While ConvNets have traditionally been the backbone for auto-encoders, recent explorations have incorporated Vision Transformers~\citep{Vaswani2017, Kolesnikov2020} (ViT) to auto-encoding. ViTVQGAN~\citep{yu2022scaling} modifies the VQGAN architecture to use a ViT and finds scaling benefits. Unified Masked Diffusion~\citep{hansenestruch2024unifiedautoencodingmaskeddiffusion} uses a ViT encoder-decoder framework for representation and generation tasks. TiTok~\citep{yu2024image} introduces a 1D tokenizer ViT that distills latent codes from VQGAN.  Finally, ElasticTok~\citep{yan2024elastictok} is concurrent work and utilizes a similar masking mechanism, though their paper focuses on reconstruction and does not try generation tasks. 

\paragraph{Video tokenization.} VideoGPT~\citep{yan2021videogpt} proposes using 3D Convolutions with a VQVAE. TATS~\cite{ge2022longvideogenerationtimeagnostic} utilizes replicate padding to reduce temporal corruptions issues with variable length videos. Phenaki~\citep{villegas2022phenaki} utilizes the Video Vision Transformer~\citep{arnab2021vivit}(ViViT) architecture with a factorized attention using full spatial and casual temporal attention. MAGViTv1~\citep{yu2023magvit, yu2023language} utilizes a 3D convolution with VQGAN to learn a video tokenizer coupled with a masked generative portion. The temporal auto-encoder (TAE) used in Movie Gen~\citep{polyak2024moviegencastmedia} is a continuous noncausal 2.5D CNN tokenizer that allows for variable resolutions and video length encodings. Finally, LARP~\citep{wang2024larp} is concurrent works that tokenizes videos with ViT into discrete codes similar to TiTok's architecture~\citep{yu2024image}, our work differs as we use continuous codes and don't concatenate latent tokens to the encoder.

\paragraph{High resolution generation.} High resolution image generation has been done prior from sampling VAEs, GANs~\citep{goodfellow2014generative}, and Diffusion Models~\citep{SohlDickstein2015,Song2019,Song2020,Ho2020}. While some work perform image synthesis in pixel space~\citep{Dhariwal2021}, many works have found it more computationally effective to perform generation in a latent space from an auto-encoder~\citep{Rombach2022}. 

Typically the U-Net architecture~\citep{ronneberger2015u} has been used for diffusion modeling, though recently transformers have been gaining favor in image generation. MaskGIT~\citep{chang2022maskgit} combines masking tokens with a schedule to generate images and Diffusion Transformers~\citep{Peebles2023} (DiT) proposes to replace the U-Net architecture with a ViT with adaptive layer normalization. Some methods use auto-regressive modeling to generate images~\citep{Ramesh2021, yu2023magvit, yu2023language, li2024autoregressive}.

DALL-E~\citep{Ramesh2021} encodes images with a VQVAE and then uses next token prediction to generate the images. While most auto-regressive image generators rely on discrete image spaces, MAR~\citep{li2024autoregressive} proposed a synergized next token predictor that allows for visual modeling in continuous latent spaces. 

\section{Conclusion}
In this paper, we explored scaling in auto-encoders. We introduced \method, a ViT-style auto-encoder to perform exploration. We tested scaling bottleneck sizes, encoder sizes, and decoder sizes. We found a strong correlation between the total number of floating points ($E$) and visual quality metrics. Our findings indicate that scaling the auto-encoder size alone does not significantly enhance downstream generative performance. Specifically, increasing the bottleneck size improves reconstruction quality but complicates training and negatively impacts generation when the latent space becomes too large. Additionally, scaling the encoder often fails to boost performance and can be detrimental, while scaling the decoder offers mixed results—enhancing reconstruction but not consistently improving generative tasks. These trends hold true for both image and video tokenizers, with our proposed \method effectively leveraging redundancy in video data to achieve superior performance in video generation tasks.

The best performing \method from our sweep achieves highly competitive performance with state-of-the-art tokenizers, matching rFID and rFVD metrics while requiring significantly fewer FLOPs. In benchmarks such as ImageNet, COCO, and UCF-101, \method not only matches but in some cases surpasses existing methods, particularly in class-conditional video generation. Our study highlights critical factors in the design and scaling of visual tokenizers, emphasizing the importance of bottleneck design and the nuanced effects of encoder and decoder scaling. We hope that our work will inspire further research into effective Transformer-based architectures for visual tokenization, ultimately advancing the field of high-quality image and video generation.

\section{Acknowledgments}
This research was conducted during an internship at Meta with compute from Meta's AWS servers. We thank Meta and all internal collaborators for their support and resources. Special thanks to Animesh Sinha; to Kaiming He and Tianhong Li for related discussions. 

\bibliographystyle{assets/plainnat}
\bibliography{main}

\begin{thebibliography}{71}
\providecommand{\natexlab}[1]{#1}
\providecommand{\url}[1]{\texttt{#1}}
\expandafter\ifx\csname urlstyle\endcsname\relax
  \providecommand{\doi}[1]{doi: #1}\else
  \providecommand{\doi}{doi: \begingroup \urlstyle{rm}\Url}\fi

\bibitem[Arnab et~al.(2021)Arnab, Dehghani, Heigold, Sun, Lu{\v{c}}i{\'c}, and Schmid]{arnab2021vivit}
Anurag Arnab, Mostafa Dehghani, Georg Heigold, Chen Sun, Mario Lu{\v{c}}i{\'c}, and Cordelia Schmid.
\newblock Vivit: A video vision transformer.
\newblock In \emph{Proceedings of the IEEE/CVF international conference on computer vision}, pages 6836--6846, 2021.

\bibitem[Ba et~al.(2016)Ba, Kiros, and Hinton]{Ba2016}
Jimmy~Lei Ba, Jamie~Ryan Kiros, and Geoffrey~E Hinton.
\newblock Layer normalization.
\newblock \emph{arXiv:1607.06450}, 2016.

\bibitem[Beyer et~al.(2022)Beyer, Zhai, and Kolesnikov]{big_vision}
Lucas Beyer, Xiaohua Zhai, and Alexander Kolesnikov.
\newblock Big vision.
\newblock \url{https://github.com/google-research/big_vision}, 2022.

\bibitem[Brooks et~al.(2024)Brooks, Peebles, Holmes, DePue, Guo, Jing, Schnurr, Taylor, Luhman, Luhman, et~al.]{brooks2024video}
Tim Brooks, Bill Peebles, Connor Holmes, Will DePue, Yufei Guo, Li~Jing, David Schnurr, Joe Taylor, Troy Luhman, Eric Luhman, et~al.
\newblock Video generation models as world simulators. 2024.
\newblock \emph{URL https://openai. com/research/video-generation-models-as-world-simulators}, 2024.

\bibitem[Chang et~al.(2022)Chang, Zhang, Jiang, Liu, and Freeman]{chang2022maskgit}
Huiwen Chang, Han Zhang, Lu~Jiang, Ce~Liu, and William~T Freeman.
\newblock Maskgit: Masked generative image transformer.
\newblock In \emph{CVPR}, 2022.

\bibitem[Deng et~al.(2009)Deng, Dong, Socher, Li, Li, and Fei-Fei]{Deng2009}
Jia Deng, Wei Dong, Richard Socher, Li-Jia Li, Kai Li, and Li~Fei-Fei.
\newblock {ImageNet: A large-scale hierarchical image database}.
\newblock In \emph{CVPR}, 2009.

\bibitem[Dhariwal and Nichol(2021)]{Dhariwal2021}
Prafulla Dhariwal and Alexander Nichol.
\newblock Diffusion models beat {GANs} on image synthesis.
\newblock In \emph{NIPS}, 2021.

\bibitem[Dosovitskiy et~al.(2021)Dosovitskiy, Beyer, Kolesnikov, Weissenborn, Zhai, Unterthiner, Dehghani, Minderer, Heigold, Gelly, Uszkoreit, and Houlsby]{Dosovitskiy2021}
Alexey Dosovitskiy, Lucas Beyer, Alexander Kolesnikov, Dirk Weissenborn, Xiaohua Zhai, Thomas Unterthiner, Mostafa Dehghani, Matthias Minderer, Georg Heigold, Sylvain Gelly, Jakob Uszkoreit, and Neil Houlsby.
\newblock An image is worth 16x16 words: Transformers for image recognition at scale.
\newblock In \emph{ICLR}, 2021.

\bibitem[Esser et~al.(2021)Esser, Rombach, and Ommer]{Esser2021}
Patrick Esser, Robin Rombach, and Bj{\"o}rn Ommer.
\newblock Taming transformers for high-resolution image synthesis.
\newblock In \emph{CVPR}, 2021.

\bibitem[Esser et~al.(2024)Esser, Kulal, Blattmann, Entezari, M{\"u}ller, Saini, Levi, Lorenz, Sauer, Boesel, et~al.]{esser2024scaling}
Patrick Esser, Sumith Kulal, Andreas Blattmann, Rahim Entezari, Jonas M{\"u}ller, Harry Saini, Yam Levi, Dominik Lorenz, Axel Sauer, Frederic Boesel, et~al.
\newblock Scaling rectified flow transformers for high-resolution image synthesis.
\newblock In \emph{ICML}, 2024.

\bibitem[Ge et~al.(2022)Ge, Hayes, Yang, Yin, Pang, Jacobs, Huang, and Parikh]{ge2022longvideogenerationtimeagnostic}
Songwei Ge, Thomas Hayes, Harry Yang, Xi~Yin, Guan Pang, David Jacobs, Jia-Bin Huang, and Devi Parikh.
\newblock Long video generation with time-agnostic vqgan and time-sensitive transformer, 2022.
\newblock \url{https://arxiv.org/abs/2204.03638}.

\bibitem[Genmo(2024)]{genmo2024mochi}
Genmo.
\newblock Mochi 1.
\newblock \url{https://github.com/genmoai/models}, 2024.

\bibitem[Goodfellow et~al.(2014)Goodfellow, Pouget-Abadie, Mirza, Xu, Warde-Farley, Ozair, Courville, and Bengio]{goodfellow2014generative}
Ian Goodfellow, Jean Pouget-Abadie, Mehdi Mirza, Bing Xu, David Warde-Farley, Sherjil Ozair, Aaron Courville, and Yoshua Bengio.
\newblock Generative adversarial nets.
\newblock \emph{Advances in neural information processing systems}, 27, 2014.

\bibitem[Gu and Dao(2023)]{gu2023mamba}
Albert Gu and Tri Dao.
\newblock Mamba: Linear-time sequence modeling with selective state spaces.
\newblock \emph{arXiv preprint arXiv:2312.00752}, 2023.

\bibitem[Hansen-Estruch et~al.(2024)Hansen-Estruch, Vishwanath, Zhang, and Tomar]{hansenestruch2024unifiedautoencodingmaskeddiffusion}
Philippe Hansen-Estruch, Sriram Vishwanath, Amy Zhang, and Manan Tomar.
\newblock Unified auto-encoding with masked diffusion, 2024.
\newblock \url{https://arxiv.org/abs/2406.17688}.

\bibitem[He et~al.(2016)He, Zhang, Ren, and Sun]{He2016}
Kaiming He, Xiangyu Zhang, Shaoqing Ren, and Jian Sun.
\newblock Deep residual learning for image recognition.
\newblock In \emph{CVPR}, 2016.

\bibitem[He et~al.(2022)He, Chen, Xie, Li, Doll{\'a}r, and Girshick]{He2022}
Kaiming He, Xinlei Chen, Saining Xie, Yanghao Li, Piotr Doll{\'a}r, and Ross Girshick.
\newblock Masked autoencoders are scalable vision learners.
\newblock In \emph{CVPR}, 2022.

\bibitem[Heusel et~al.(2017)Heusel, Ramsauer, Unterthiner, Nessler, and Hochreiter]{Heusel2017}
Martin Heusel, Hubert Ramsauer, Thomas Unterthiner, Bernhard Nessler, and Sepp Hochreiter.
\newblock {GANs} trained by a two time-scale update rule converge to a local {Nash} equilibrium.
\newblock In \emph{NIPS}, 2017.

\bibitem[Hinton et~al.(2012)Hinton, Srivastava, Krizhevsky, Sutskever, and Salakhutdinov]{Hinton2012}
Geoffrey~E Hinton, Nitish Srivastava, Alex Krizhevsky, Ilya Sutskever, and Ruslan~R Salakhutdinov.
\newblock Improving neural networks by preventing co-adaptation of feature detectors.
\newblock \emph{arXiv:1207.0580}, 2012.

\bibitem[Ho and Salimans(2022)]{ho2022classifier}
Jonathan Ho and Tim Salimans.
\newblock Classifier-free diffusion guidance.
\newblock \emph{arXiv preprint arXiv:2207.12598}, 2022.

\bibitem[Ho et~al.(2020)Ho, Jain, and Abbeel]{Ho2020}
Jonathan Ho, Ajay Jain, and Pieter Abbeel.
\newblock Denoising diffusion probabilistic models.
\newblock In \emph{NIPS}, 2020.

\bibitem[Johnson et~al.(2016)Johnson, Alahi, and Fei-Fei]{johnson2016perceptual}
Justin Johnson, Alexandre Alahi, and Li~Fei-Fei.
\newblock Perceptual losses for real-time style transfer and super-resolution.
\newblock In \emph{Computer Vision--ECCV 2016: 14th European Conference, Amsterdam, The Netherlands, October 11-14, 2016, Proceedings, Part II 14}, pages 694--711. Springer, 2016.

\bibitem[Karras et~al.(2019)Karras, Laine, and Aila]{karras2019stylebasedgeneratorarchitecturegenerative}
Tero Karras, Samuli Laine, and Timo Aila.
\newblock A style-based generator architecture for generative adversarial networks, 2019.
\newblock \url{https://arxiv.org/abs/1812.04948}.

\bibitem[Kay et~al.(2017)Kay, Carreira, Simonyan, Zhang, Hillier, Vijayanarasimhan, Viola, Green, Back, Natsev, et~al.]{Kay2017}
Will Kay, Joao Carreira, Karen Simonyan, Brian Zhang, Chloe Hillier, Sudheendra Vijayanarasimhan, Fabio Viola, Tim Green, Trevor Back, Paul Natsev, et~al.
\newblock The {Kinetics} human action video dataset.
\newblock \emph{arXiv:1705.06950}, 2017.

\bibitem[Kingma and Ba(2015)]{Kingma2015}
Diederik~P Kingma and Jimmy Ba.
\newblock Adam: A method for stochastic optimization.
\newblock In \emph{ICLR}, 2015.

\bibitem[Kingma and Welling(2013)]{Kingma2013}
Diederik~P Kingma and Max Welling.
\newblock Auto-encoding variational bayes.
\newblock In \emph{ICLR}, 2013.

\bibitem[Kolesnikov et~al.(2020)Kolesnikov, Beyer, Zhai, Puigcerver, Yung, Gelly, and Houlsby]{Kolesnikov2020}
Alexander Kolesnikov, Lucas Beyer, Xiaohua Zhai, Joan Puigcerver, Jessica Yung, Sylvain Gelly, and Neil Houlsby.
\newblock {Big Transfer (BiT)}: General visual representation learning.
\newblock In \emph{ECCV}, 2020.

\bibitem[LeCun et~al.(1998)LeCun, Bottou, Bengio, Haffner, et~al.]{lecun1998gradient}
Yann LeCun, L{\'e}on Bottou, Yoshua Bengio, Patrick Haffner, et~al.
\newblock Gradient-based learning applied to document recognition.
\newblock \emph{Proceedings of the IEEE}, 86\penalty0 (11):\penalty0 2278--2324, 1998.

\bibitem[Lee et~al.(2022)Lee, Kim, Kim, Cho, and Han]{lee2022autoregressiveimagegenerationusing}
Doyup Lee, Chiheon Kim, Saehoon Kim, Minsu Cho, and Wook-Shin Han.
\newblock Autoregressive image generation using residual quantization, 2022.
\newblock \url{https://arxiv.org/abs/2203.01941}.

\bibitem[Li et~al.(2024)Li, Tian, Li, Deng, and He]{li2024autoregressive}
Tianhong Li, Yonglong Tian, He~Li, Mingyang Deng, and Kaiming He.
\newblock Autoregressive image generation without vector quantization.
\newblock \emph{arXiv preprint arXiv:2406.11838}, 2024.

\bibitem[Lin et~al.(2014{\natexlab{a}})Lin, Chen, and Yan]{Lin2014a}
Min Lin, Qiang Chen, and Shuicheng Yan.
\newblock Network in network.
\newblock In \emph{ICLR}, 2014{\natexlab{a}}.

\bibitem[Lin et~al.(2014{\natexlab{b}})Lin, Maire, Belongie, Hays, Perona, Ramanan, Doll{\'a}r, and Zitnick]{Lin2014}
Tsung-Yi Lin, Michael Maire, Serge Belongie, James Hays, Pietro Perona, Deva Ramanan, Piotr Doll{\'a}r, and C~Lawrence Zitnick.
\newblock {Microsoft COCO: Common objects in context}.
\newblock In \emph{ECCV}, 2014{\natexlab{b}}.

\bibitem[Loshchilov(2017)]{loshchilov2017decoupled}
I~Loshchilov.
\newblock Decoupled weight decay regularization.
\newblock \emph{arXiv preprint arXiv:1711.05101}, 2017.

\bibitem[Loshchilov and Hutter(2017)]{Loshchilov2016}
Ilya Loshchilov and Frank Hutter.
\newblock {SGDR}: Stochastic gradient descent with warm restarts.
\newblock In \emph{ICLR}, 2017.

\bibitem[Mentzer et~al.(2023)Mentzer, Minnen, Agustsson, and Tschannen]{mentzer2023finite}
Fabian Mentzer, David Minnen, Eirikur Agustsson, and Michael Tschannen.
\newblock Finite scalar quantization: Vq-vae made simple.
\newblock \emph{arXiv preprint arXiv:2309.15505}, 2023.

\bibitem[NVIDIA(2024)]{nvidia_cosmos_tokenizer}
NVIDIA.
\newblock Cosmos-tokenizer.
\newblock \url{https://github.com/NVIDIA/Cosmos-Tokenizer}, 2024.
\newblock Accessed: 2025-01-05. A tokenizer designed for efficient processing of large-scale datasets.

\bibitem[Oord et~al.(2017)Oord, Vinyals, and Kavukcuoglu]{Oord2017}
Aaron van~den Oord, Oriol Vinyals, and Koray Kavukcuoglu.
\newblock Neural discrete representation learning.
\newblock In \emph{NIPS}, 2017.

\bibitem[OpenAI(2023)]{consistencydecoder}
OpenAI.
\newblock Consistency decoder.
\newblock \url{https://github.com/openai/consistencydecoder}, 2023.
\newblock Accessed: 2023-10-17.

\bibitem[Paszke et~al.(2019)Paszke, Gross, Massa, Lerer, Bradbury, Chanan, Killeen, Lin, Gimelshein, Antiga, et~al.]{Paszke2019}
Adam Paszke, Sam Gross, Francisco Massa, Adam Lerer, James Bradbury, Gregory Chanan, Trevor Killeen, Zeming Lin, Natalia Gimelshein, Luca Antiga, et~al.
\newblock {PyTorch: An imperative style, high-performance deep learning library}.
\newblock In \emph{NeurIPS}, 2019.

\bibitem[Peebles and Xie(2023)]{Peebles2023}
William Peebles and Saining Xie.
\newblock Scalable diffusion models with {Transformers}.
\newblock In \emph{ICCV}, 2023.

\bibitem[Podell et~al.(2023)Podell, English, Lacey, Blattmann, Dockhorn, M{\"u}ller, Penna, and Rombach]{podell2023sdxl}
Dustin Podell, Zion English, Kyle Lacey, Andreas Blattmann, Tim Dockhorn, Jonas M{\"u}ller, Joe Penna, and Robin Rombach.
\newblock Sdxl: Improving latent diffusion models for high-resolution image synthesis.
\newblock \emph{arXiv preprint arXiv:2307.01952}, 2023.

\bibitem[Polyak et~al.(2024)Polyak, Zohar, Brown, Tjandra, Sinha, Lee, Vyas, Shi, Ma, Chuang, Yan, Choudhary, Wang, Sethi, Pang, Ma, Misra, Hou, Wang, Jagadeesh, Li, Zhang, Singh, Williamson, Le, Yu, Singh, Zhang, Vajda, Duval, Girdhar, Sumbaly, Rambhatla, Tsai, Azadi, Datta, Chen, Bell, Ramaswamy, Sheynin, Bhattacharya, Motwani, Xu, Li, Hou, Hsu, Yin, Dai, Taigman, Luo, Liu, Wu, Zhao, Kirstain, He, He, Pumarola, Thabet, Sanakoyeu, Mallya, Guo, Araya, Kerr, Wood, Liu, Peng, Vengertsev, Schonfeld, Blanchard, Juefei-Xu, Nord, Liang, Hoffman, Kohler, Fire, Sivakumar, Chen, Yu, Gao, Georgopoulos, Moritz, Sampson, Li, Parmeggiani, Fine, Fowler, Petrovic, and Du]{polyak2024moviegencastmedia}
Adam Polyak, Amit Zohar, Andrew Brown, Andros Tjandra, Animesh Sinha, Ann Lee, Apoorv Vyas, Bowen Shi, Chih-Yao Ma, Ching-Yao Chuang, David Yan, Dhruv Choudhary, Dingkang Wang, Geet Sethi, Guan Pang, Haoyu Ma, Ishan Misra, Ji~Hou, Jialiang Wang, Kiran Jagadeesh, Kunpeng Li, Luxin Zhang, Mannat Singh, Mary Williamson, Matt Le, Matthew Yu, Mitesh~Kumar Singh, Peizhao Zhang, Peter Vajda, Quentin Duval, Rohit Girdhar, Roshan Sumbaly, Sai~Saketh Rambhatla, Sam Tsai, Samaneh Azadi, Samyak Datta, Sanyuan Chen, Sean Bell, Sharadh Ramaswamy, Shelly Sheynin, Siddharth Bhattacharya, Simran Motwani, Tao Xu, Tianhe Li, Tingbo Hou, Wei-Ning Hsu, Xi~Yin, Xiaoliang Dai, Yaniv Taigman, Yaqiao Luo, Yen-Cheng Liu, Yi-Chiao Wu, Yue Zhao, Yuval Kirstain, Zecheng He, Zijian He, Albert Pumarola, Ali Thabet, Artsiom Sanakoyeu, Arun Mallya, Baishan Guo, Boris Araya, Breena Kerr, Carleigh Wood, Ce~Liu, Cen Peng, Dimitry Vengertsev, Edgar Schonfeld, Elliot Blanchard, Felix Juefei-Xu, Fraylie Nord, Jeff Liang, John Hoffman, Jonas
  Kohler, Kaolin Fire, Karthik Sivakumar, Lawrence Chen, Licheng Yu, Luya Gao, Markos Georgopoulos, Rashel Moritz, Sara~K. Sampson, Shikai Li, Simone Parmeggiani, Steve Fine, Tara Fowler, Vladan Petrovic, and Yuming Du.
\newblock Movie gen: A cast of media foundation models, 2024.
\newblock \url{https://arxiv.org/abs/2410.13720}.

\bibitem[Ramesh et~al.(2021)Ramesh, Pavlov, Goh, Gray, Voss, Radford, Chen, and Sutskever]{Ramesh2021}
Aditya Ramesh, Mikhail Pavlov, Gabriel Goh, Scott Gray, Chelsea Voss, Alec Radford, Mark Chen, and Ilya Sutskever.
\newblock Zero-shot text-to-image generation.
\newblock In \emph{ICML}, 2021.

\bibitem[Rombach et~al.(2022)Rombach, Blattmann, Lorenz, Esser, and Ommer]{Rombach2022}
Robin Rombach, Andreas Blattmann, Dominik Lorenz, Patrick Esser, and Bj{\"o}rn Ommer.
\newblock High-resolution image synthesis with latent diffusion models.
\newblock In \emph{CVPR}, 2022.

\bibitem[Ronneberger et~al.(2015)Ronneberger, Fischer, and Brox]{ronneberger2015u}
Olaf Ronneberger, Philipp Fischer, and Thomas Brox.
\newblock U-net: Convolutional networks for biomedical image segmentation.
\newblock In \emph{MICCAI}, 2015.

\bibitem[Salimans et~al.(2016)Salimans, Goodfellow, Zaremba, Cheung, Radford, and Chen]{salimans2016improved}
Tim Salimans, Ian Goodfellow, Wojciech Zaremba, Vicki Cheung, Alec Radford, and Xi~Chen.
\newblock Improved techniques for training gans.
\newblock \emph{NeurIPS}, 2016.

\bibitem[Shazeer(2020)]{shazeer2020glu}
Noam Shazeer.
\newblock Glu variants improve transformer.
\newblock \emph{arXiv preprint arXiv:2002.05202}, 2020.

\bibitem[Sohl-Dickstein et~al.(2015)Sohl-Dickstein, Weiss, Maheswaranathan, and Ganguli]{SohlDickstein2015}
Jascha Sohl-Dickstein, Eric Weiss, Niru Maheswaranathan, and Surya Ganguli.
\newblock Deep unsupervised learning using nonequilibrium thermodynamics.
\newblock In \emph{ICML}, 2015.

\bibitem[Song and Ermon(2019)]{Song2019}
Yang Song and Stefano Ermon.
\newblock Generative modeling by estimating gradients of the data distribution.
\newblock \emph{NIPS}, 2019.

\bibitem[Song et~al.(2020)Song, Sohl-Dickstein, Kingma, Kumar, Ermon, and Poole]{Song2020}
Yang Song, Jascha Sohl-Dickstein, Diederik~P Kingma, Abhishek Kumar, Stefano Ermon, and Ben Poole.
\newblock Score-based generative modeling through stochastic differential equations.
\newblock \emph{arXiv:2011.13456}, 2020.

\bibitem[Soomro(2012)]{soomro2012ucf101}
K~Soomro.
\newblock Ucf101: A dataset of 101 human actions classes from videos in the wild.
\newblock \emph{arXiv preprint arXiv:1212.0402}, 2012.

\bibitem[Su et~al.(2024)Su, Ahmed, Lu, Pan, Bo, and Liu]{su2024roformer}
Jianlin Su, Murtadha Ahmed, Yu~Lu, Shengfeng Pan, Wen Bo, and Yunfeng Liu.
\newblock Roformer: Enhanced transformer with rotary position embedding.
\newblock \emph{Neurocomputing}, 568:\penalty0 127063, 2024.

\bibitem[Sun et~al.(2024)Sun, Jiang, Chen, Zhang, Peng, Luo, and Yuan]{sun2024autoregressive}
Peize Sun, Yi~Jiang, Shoufa Chen, Shilong Zhang, Bingyue Peng, Ping Luo, and Zehuan Yuan.
\newblock Autoregressive model beats diffusion: Llama for scalable image generation.
\newblock \emph{arXiv preprint arXiv:2406.06525}, 2024.

\bibitem[Tomar et~al.(2024)Tomar, Hansen-Estruch, Bachman, Lamb, Langford, Taylor, and Levine]{tomar2024videooccupancymodels}
Manan Tomar, Philippe Hansen-Estruch, Philip Bachman, Alex Lamb, John Langford, Matthew~E. Taylor, and Sergey Levine.
\newblock Video occupancy models, 2024.
\newblock \url{https://arxiv.org/abs/2407.09533}.

\bibitem[Touvron et~al.(2023)Touvron, Lavril, Izacard, Martinet, Lachaux, Lacroix, Rozi{\`e}re, Goyal, Hambro, Azhar, et~al.]{touvron2023llama}
Hugo Touvron, Thibaut Lavril, Gautier Izacard, Xavier Martinet, Marie-Anne Lachaux, Timoth{\'e}e Lacroix, Baptiste Rozi{\`e}re, Naman Goyal, Eric Hambro, Faisal Azhar, et~al.
\newblock Llama: Open and efficient foundation language models.
\newblock \emph{arXiv preprint arXiv:2302.13971}, 2023.

\bibitem[Unterthiner et~al.(2019)Unterthiner, van Steenkiste, Kurach, Marinier, Michalski, and Gelly]{unterthiner2019accurategenerativemodelsvideo}
Thomas Unterthiner, Sjoerd van Steenkiste, Karol Kurach, Raphael Marinier, Marcin Michalski, and Sylvain Gelly.
\newblock Towards accurate generative models of video: A new metric \& challenges, 2019.
\newblock \url{https://arxiv.org/abs/1812.01717}.

\bibitem[Vaswani et~al.(2017)Vaswani, Shazeer, Parmar, Uszkoreit, Jones, Gomez, Kaiser, and Polosukhin]{Vaswani2017}
Ashish Vaswani, Noam Shazeer, Niki Parmar, Jakob Uszkoreit, Llion Jones, Aidan~N Gomez, Lukasz Kaiser, and Illia Polosukhin.
\newblock Attention is all you need.
\newblock In \emph{NeurIPS}, 2017.

\bibitem[Villegas et~al.(2022)Villegas, Babaeizadeh, Kindermans, Moraldo, Zhang, Saffar, Castro, Kunze, and Erhan]{villegas2022phenaki}
Ruben Villegas, Mohammad Babaeizadeh, Pieter-Jan Kindermans, Hernan Moraldo, Han Zhang, Mohammad~Taghi Saffar, Santiago Castro, Julius Kunze, and Dumitru Erhan.
\newblock Phenaki: Variable length video generation from open domain textual descriptions.
\newblock In \emph{International Conference on Learning Representations}, 2022.

\bibitem[Vincent et~al.(2008)Vincent, Larochelle, Bengio, and Manzagol]{Vincent2008}
Pascal Vincent, Hugo Larochelle, Yoshua Bengio, and Pierre-Antoine Manzagol.
\newblock Extracting and composing robust features with denoising autoencoders.
\newblock In \emph{ICML}, 2008.

\bibitem[Wang et~al.(2024)Wang, Suri, Ren, Chen, and Shrivastava]{wang2024larp}
Hanyu Wang, Saksham Suri, Yixuan Ren, Hao Chen, and Abhinav Shrivastava.
\newblock Larp: Tokenizing videos with a learned autoregressive generative prior.
\newblock \emph{arXiv preprint arXiv:2410.21264}, 2024.

\bibitem[Wang et~al.(2023)Wang, Huang, Zhao, Tong, He, Wang, Wang, and Qiao]{wang2023videomae}
Limin Wang, Bingkun Huang, Zhiyu Zhao, Zhan Tong, Yinan He, Yi~Wang, Yali Wang, and Yu~Qiao.
\newblock Videomae v2: Scaling video masked autoencoders with dual masking.
\newblock In \emph{Proceedings of the IEEE/CVF Conference on Computer Vision and Pattern Recognition}, pages 14549--14560, 2023.

\bibitem[Wang et~al.(2004)Wang, Bovik, Sheikh, and Simoncelli]{1284395}
Zhou Wang, A.C. Bovik, H.R. Sheikh, and E.P. Simoncelli.
\newblock Image quality assessment: from error visibility to structural similarity.
\newblock \emph{IEEE Transactions on Image Processing}, 13\penalty0 (4):\penalty0 600--612, 2004.
\newblock \doi{10.1109/TIP.2003.819861}.

\bibitem[Yan et~al.(2021)Yan, Zhang, Abbeel, and Srinivas]{yan2021videogpt}
Wilson Yan, Yunzhi Zhang, Pieter Abbeel, and Aravind Srinivas.
\newblock Videogpt: Video generation using vq-vae and transformers.
\newblock \emph{arXiv preprint arXiv:2104.10157}, 2021.

\bibitem[Yan et~al.(2024)Yan, Zaharia, Mnih, Abbeel, Faust, and Liu]{yan2024elastictok}
Wilson Yan, Matei Zaharia, Volodymyr Mnih, Pieter Abbeel, Aleksandra Faust, and Hao Liu.
\newblock Elastictok: Adaptive tokenization for image and video.
\newblock \emph{arXiv preprint arXiv:2410.08368}, 2024.

\bibitem[Yao and Wang(2025)]{yao2025reconstructionvsgenerationtaming}
Jingfeng Yao and Xinggang Wang.
\newblock Reconstruction vs. generation: Taming optimization dilemma in latent diffusion models, 2025.
\newblock \url{https://arxiv.org/abs/2501.01423}.

\bibitem[Yu et~al.(2021)Yu, Li, Koh, Zhang, Pang, Qin, Ku, Xu, Baldridge, and Wu]{yu2021vector}
Jiahui Yu, Xin Li, Jing~Yu Koh, Han Zhang, Ruoming Pang, James Qin, Alexander Ku, Yuanzhong Xu, Jason Baldridge, and Yonghui Wu.
\newblock Vector-quantized image modeling with improved vqgan.
\newblock \emph{arXiv preprint arXiv:2110.04627}, 2021.

\bibitem[Yu et~al.(2022)Yu, Xu, Koh, Luong, Baid, Wang, Vasudevan, Ku, Yang, Ayan, et~al.]{yu2022scaling}
Jiahui Yu, Yuanzhong Xu, Jing~Yu Koh, Thang Luong, Gunjan Baid, Zirui Wang, Vijay Vasudevan, Alexander Ku, Yinfei Yang, Burcu~Karagol Ayan, et~al.
\newblock Scaling autoregressive models for content-rich text-to-image generation.
\newblock \emph{arXiv preprint arXiv:2206.10789}, 2\penalty0 (3):\penalty0 5, 2022.

\bibitem[Yu et~al.(2023{\natexlab{a}})Yu, Cheng, Sohn, Lezama, Zhang, Chang, Hauptmann, Yang, Hao, Essa, et~al.]{yu2023magvit}
Lijun Yu, Yong Cheng, Kihyuk Sohn, Jos{\'e} Lezama, Han Zhang, Huiwen Chang, Alexander~G Hauptmann, Ming-Hsuan Yang, Yuan Hao, Irfan Essa, et~al.
\newblock Magvit: Masked generative video transformer.
\newblock In \emph{Proceedings of the IEEE/CVF Conference on Computer Vision and Pattern Recognition}, pages 10459--10469, 2023{\natexlab{a}}.

\bibitem[Yu et~al.(2023{\natexlab{b}})Yu, Lezama, Gundavarapu, Versari, Sohn, Minnen, Cheng, Gupta, Gu, Hauptmann, et~al.]{yu2023language}
Lijun Yu, Jos{\'e} Lezama, Nitesh~B Gundavarapu, Luca Versari, Kihyuk Sohn, David Minnen, Yong Cheng, Agrim Gupta, Xiuye Gu, Alexander~G Hauptmann, et~al.
\newblock Language model beats diffusion--tokenizer is key to visual generation.
\newblock \emph{arXiv preprint arXiv:2310.05737}, 2023{\natexlab{b}}.

\bibitem[Yu et~al.(2024)Yu, Weber, Deng, Shen, Cremers, and Chen]{yu2024image}
Qihang Yu, Mark Weber, Xueqing Deng, Xiaohui Shen, Daniel Cremers, and Liang-Chieh Chen.
\newblock An image is worth 32 tokens for reconstruction and generation.
\newblock \emph{arXiv preprint arXiv:2406.07550}, 2024.

\bibitem[Zohar et~al.(2024)Zohar, Wang, Dubois, Mehta, Xiao, Hansen-Estruch, Yu, Wang, Juefei-Xu, Zhang, et~al.]{zohar2024apollo}
Orr Zohar, Xiaohan Wang, Yann Dubois, Nikhil Mehta, Tong Xiao, Philippe Hansen-Estruch, Licheng Yu, Xiaofang Wang, Felix Juefei-Xu, Ning Zhang, et~al.
\newblock Apollo: An exploration of video understanding in large multimodal models.
\newblock \emph{arXiv preprint arXiv:2412.10360}, 2024.

\end{thebibliography}

\clearpage
\beginappendix

In the appendix section we include more details on experiments, architecture details, and visualizations. 

We provide additional details on the implementaiton of \method. Our implementation is based on the VideoMAEv2~\citep{wang2023videomae} codebase and inspired by the Big Vision codebase~\citep{big_vision}. Utilizing PyTorch~\citep{Paszke2019}, we employ Distributed Data Parallel (DDP) for efficient multi-GPU training, along with activation checkpointing, bfloat16 precision, and Torch Compile optimizations. For image models, we train using 8 NVIDIA H100 GPUs, where \method S-B/16 requires approximately 6–12 hours for stage 1 and 3–6 hours for stage 2 on 256p and 512p resolutions. In comparison, DiT image models take around 72–96 hours to train for 4 million steps on the same resolutions. For video models, \method S-B/4x8 is trained on 16 NVIDIA H100 GPUs, taking about 24 hours for stage 1 and 12 hours for stage 2 on 256p, 16-frame videos, and 12 hours for 128p, 16-frame videos. DiT video models require roughly 48–96 hours to train for 500k steps with a batch size of 256. Our transformer architecture is based on the Vision Transformer (ViT)~\citep{Dosovitskiy2021} and modified to incorporate elements from the Llama architecture, including SwiGLU~\citep{shazeer2020glu} activation functions and 3D axial Rotary Position Embeddings (RoPE)~\citep{su2024roformer}. The architecture consists of Transformer blocks~\citep{Vaswani2017} with multi-head self-attention and MLP layers, enhanced by residual connections~\citep{He2016} and layer normalization~\citep{Ba2016}, closely following the Masked Autoencoder (MAE) design~\citep{He2022}. Additionally, we integrate video processing code from Apollo~\citep{zohar2024apollo} and Video Occupancy Models~\citep{tomar2024videooccupancymodels}, enabling \method to effectively handle and exploit redundancy in video data, thereby improving both reconstruction metrics and compression efficiency. Overall, \method leverages advanced training techniques and architectural innovations to achieve state-of-the-art performance in image and video reconstruction and generation tasks.

\section{Extra Experiments}\label{sec:additional_experiments}
\subsection{Detailed 256p Image Results}

\begin{figure}[h]
    \centering
    \makebox[\textwidth][c]{
        \includegraphics[width=1.0\textwidth]{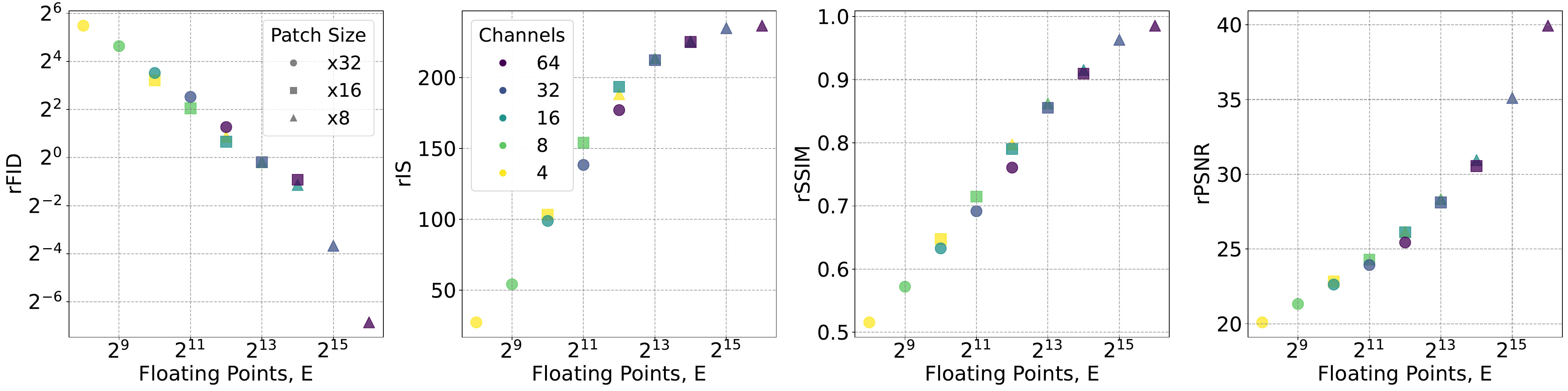}
    }
    \caption{\textbf{256p Detailed Image Reconstruction Results with Fixed Architecture Size.} We provide more details for the sweep in Figure~\ref{fig:256p_image_sweep} on the just the ImageNet-1K validation set. For $1024 \leq E \leq 16384$, where intersections of $E$ exist across patch sizes, we see very little variation in performance for fixed $E$. This indicates that $E$ is the main bottleneck for visual auto-encoding and is not influence by increasing FLOPs.}
    \label{fig:256p_detailed}
\end{figure}

We provide further detail of the ImageNet-1K validation reconstruction results from Figure~\ref{fig:256p_image_sweep} in Figure~\ref{fig:256p_detailed}. Here we show different patch sizes and channels over $E$. This shows that regardless of patch size and FLOPs usage, $E$ is highly correlated with the reconstruction perforance

\subsection{GAN Fine-tuning Ablation}

\begin{figure}[h]
\centering
\makebox[\textwidth][c]{
    \includegraphics[width=1.0\textwidth]{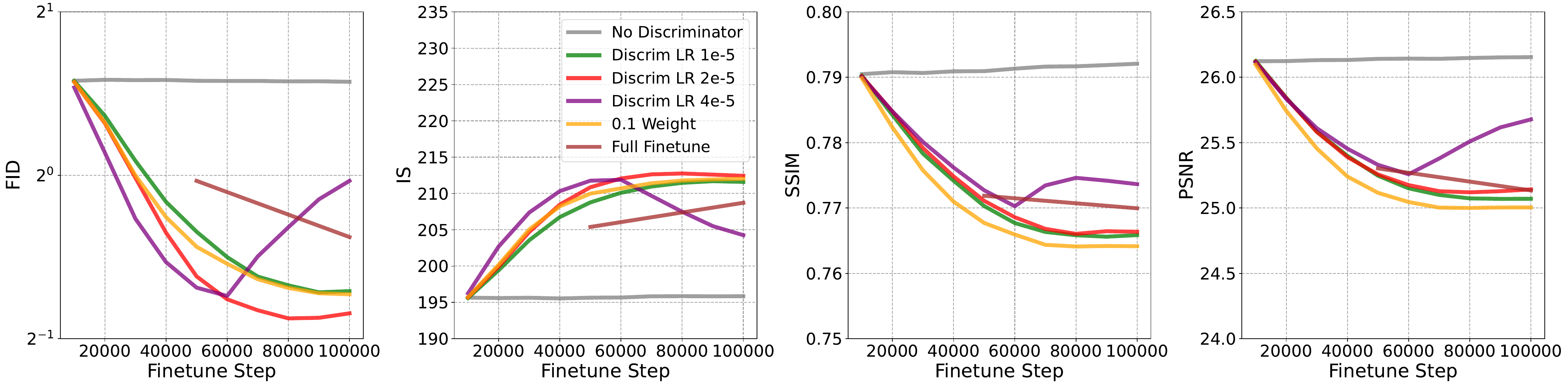}
}
\caption{\textbf{Finetuning the Decoder with a GAN.} We study the effects of finetuning the decoder in \method S-B/16 on 256p images. We compare: (1) no GAN finetuning, (2) different discriminator learning rates, (3) an increased GAN loss weight (0.1), and (4) a full finetuning of all model parameters (including the encoder). The best results occur with a discriminator learning rate of \(2\times10^{-5}\), while higher rates cause instabilities. We also observe a clear shift toward more generative behavior: as the decoder gains better IS/FID, it sacrifices some SSIM/PSNR, reflecting its transition into a stronger generative component.
}
\label{fig:256p_image_finetune_steps_vs_metrics_gan}
\end{figure}

In Figure~\ref{fig:256p_image_finetune_steps_vs_metrics_gan}, we study how various loss settings affect finetuning of the GAN decoder. Our goal is to highlight the trade-off and the decoder’s transition toward more generative behavior. We use \method S-B/16 on 256p images, following the protocol in Section~\ref{sec:Experimental_Setup} for stage 2 fine-tuning from a model trained on stage 1.

We compare:
\begin{itemize}
    \item Finetuning the decoder with the same Stage~1 loss (no GAN).
    \item Finetuning with discriminator learning rates (\(\{1\times10^{-5}, 2\times10^{-5}, 4\times10^{-5}\}\)) and a GAN weight of 0.05.
    \item Finetuning the full encoder/decoder with the GAN.
    \item Using a higher GAN weight of 0.1 with a discriminator learning rate of \(1\times10^{-5}\).
\end{itemize}

From Figure~\ref{fig:256p_image_finetune_steps_vs_metrics_gan}, the best setting is a GAN weight of 0.05 and a discriminator learning rate of \(2\times10^{-5}\). A higher discriminator learning rate causes training instabilities, while a lower rate degrades performance. Full finetuning works but does slightly worse than just finetuning the decoder. Finetuning without a GAN shows no improvement, confirming that GAN training is the primary driver of better results.

Finally, we see an inherent trade-off: improving FID tends to worsen SSIM/PSNR, indicating that as the decoder focuses on visual fidelity, it shifts more toward generative outputs. This demonstrates the decoder’s evolving role as a generative model to enhance visual performance.

\subsection{Latent \method and Masked ViTok}\label{sec:latent_mask_vitok}
In this section, we describe two variants of \method that provide different potential directions for tokenization. First we describe and evaluate our latent variation that does 1D tokenization and can form more arbitrary code shapes, then we discuss and evaluate our masking variant that allows for variable, adaptive tokenization.

\paragraph{\textbf{Latent \method Variation}.} Another variant of \method involves utilizing latent codes following Titok~\citep{yu2024image}. Initially, after applying a tubelet embedding, we concatenate a set of 1D sincos initialized latent tokens with dimensions $l_{\text{latent}} \times C_f$ to the tubelet token sequence $X_{\text{embed}}$. This combined sequence is then processed through the encoder and bottleneck using a linear layer. Subsequently, the tubelet tokens are discarded, and the latent tokens output by the encoder form $Z = l_{\text{latent}} \times 2c$, from which we sample $z \sim Z$. This gives us a 1D code with easy shape manipulation since $L$ and $c$ is arbitrarly decided and not dependent on $p$. In the decoder, $z$ is upsampled to $C_g$, and we concatenate a flattened masked token sequence of length $L\times C_g$ with the upsampled latent code $l_{\text{latent}} \times C_g$. The decoder then predicts $\hat{X}$ in the same manner as the simple \method variation using the masked tokens. This approach allows for a more adaptive compression size and shape using self attention. Additionally, it accommodates arbitrary code shapes of different lengths than $L$, provided there is redundancy in the code. A trade-off compared to the simple \method is the increased total sequence length and computational cost (FLOPs) during encoding and decoding. We refer to this variant as Latent \method.

\begin{figure}[h]
\centering
\makebox[\textwidth][c]{
    \includegraphics[width=1.0\textwidth]{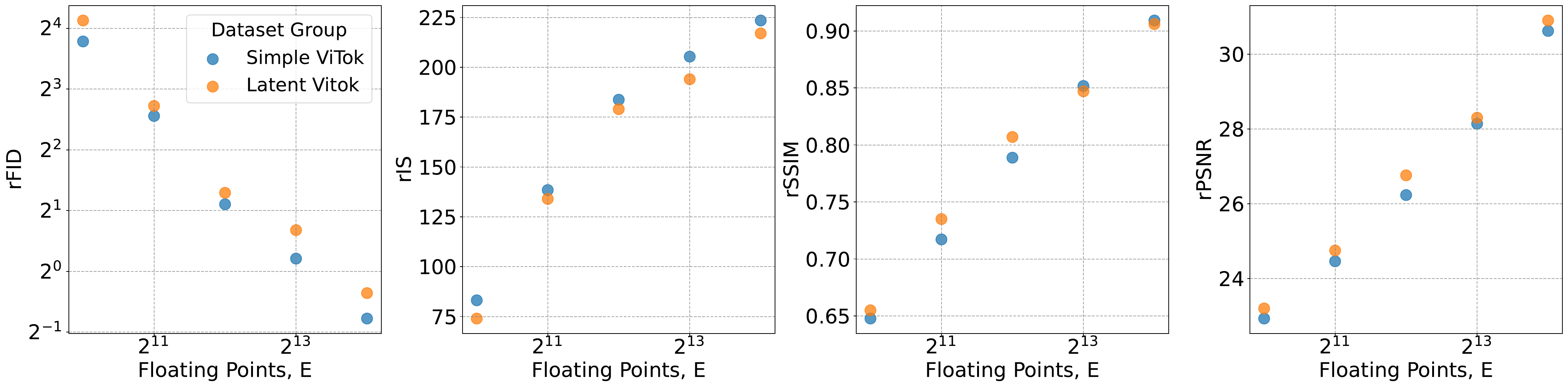}
}
\caption{\textbf{256p Simple vs Latent \method Results.} We implement a latent variant of \method S-B/16, with $p=16$ and $L \in \{64, 128, 256, 512, 1024\}$ latent tokens appended to the original patch embedding, then processed using full self-attention, and subsequently bottlenecked to $c=16$. Although this latent variant slightly underperforms the simpler version in rFID/rIS, it remains comparable overall and follows the same rules as $E$. Consequently, it provides an alternative to Simple \method with greater control over the latent space.}
\label{fig:256p_latent_vs_simple}
\end{figure}

We train latent \method on stage 1 (Section~\ref{sec:Experimental_Setup}) where we fix $c=16$ and sweep the number of latent tokens $L \in \{64, 128, 256, 512, 1024\}$ to adjust $E$. The results are in Figure~\ref{fig:256p_latent_vs_simple}. Our simple variant outperforms the latent version for most values of $E$, although the latent version achieves slightly better rSSIM/rPSNR for certain choices of $E$. This indicates that the latent approach is a promising alternative to simple \method for more control over the latent space, but comes with an increased computational cost due to the longer sequence of concatenated tokens. We leave this implementation out of \method due to added complexity. 

\paragraph{\textbf{Token Compression via Random Masking.}} The simplest bottlenecking process in \method involves manipulating $c$, which does not compress the number of tokens; the token count remains equivalent to the number tokens post-patching ($L$) or equivalent to the number of latent tokens ($l_{\text{latent}}$). Though, manipulating $p$ does not provide a fine grain control over the token count.

To form another bottleneck, we can instead manipulate the main sequence of patch tokens by masking a random power of two number of tokens, starting with tokens \textit{at the end} of the sequence and masking towards the beginning.  This is similar to the method done in ElasticTok~\citep{yan2024elastictok}. For example, if we randomly select 256 as the masking amount for a sequence of 1024 tokens, then the last 256 tokens will be masked out and replaced with a learned masked token of dimension $c$. This directional masking strategy enforces an ordered structure to the tokens. We set the minimum length to $l$. The length of the code at inference, $l_{\text{eval}}$, forms another axis to change code shape (Section~\ref{sec:Exploration}) and $E = l_{\text{eval}} \times 2c$.

\begin{figure}[h]
\centering
\makebox[\textwidth][c]{
    \includegraphics[width=1.0\textwidth]{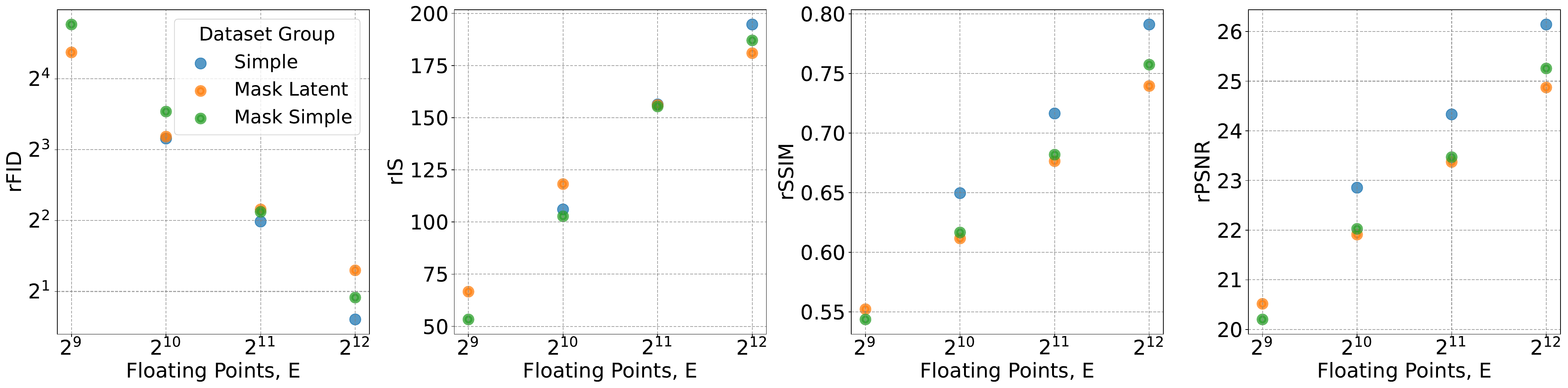}
}
\caption{\textbf{256p Adaptive Masking \method Results.} We investigate variations of \method S-B/16 that apply token masking after encoding. We consider two approaches: \emph{Mask Simple}, which masks the patch tokens following encoding, and \emph{Mask Latent}, which introduces latent tokens (like the architecture used for Figure~\ref{fig:256p_latent_vs_simple}) and masks them. At stage 1 training time we randomly selected token lengths $\{32, 64, 128, 256\}$ with $c=16$, then at inference evaluate each model on every token length and compare to the simple \method baseline at similar $E$. While the masking variations underperform the simple variant, they still perform strongly. \emph{Mask Simple} tends to perform better at higher $E$, while \emph{Mask Latent} achieves better results at lower $E$.
 }
\label{fig:256p_mask_vs_simple_latent}
\end{figure}

We now train our mask \method on stage 1 (Section~\ref{sec:Experimental_Setup}) and investigate potential adaptive tokenization schemes. We first apply this masking strategy to the simple version of \method, directly masking the patch tokens after they have been processed by the encoder. We then explore the same approach on the latent version of \method. Both methods are trained with token lengths $\{32, 64, 128, 256\}$ and $c=16$ on \method S-B/16 using 256p images.

Figure~\ref{fig:256p_mask_vs_simple_latent} compares these masking methods to the simple \method across different $E$. While all masking variants slightly underperform the simple \method, their overall performance remains strong. In particular, masking patches directly is more effective for higher $E>4096$, whereas masking latent tokens performs better when $E<4096$. These findings highlight how \method can be adapted for flexible token lengths during inference, and illustrate how our method can be extended to learn an ordered structure of tokens. Though more work here is needed to improve performance further.
\clearpage
\section{Visualizations}\label{sec:visualizations}
In this section we provide extra visualizations of generation examples from our various models and sweeps.

\subsection{Video Generations} We include more video generation results in this section from Table~\ref{tab:video_gen_results} and show example generations at 512 and 256 tokens respectively. 

\begin{figure}[h]
    \centering
    \includegraphics[width=0.95\textwidth]{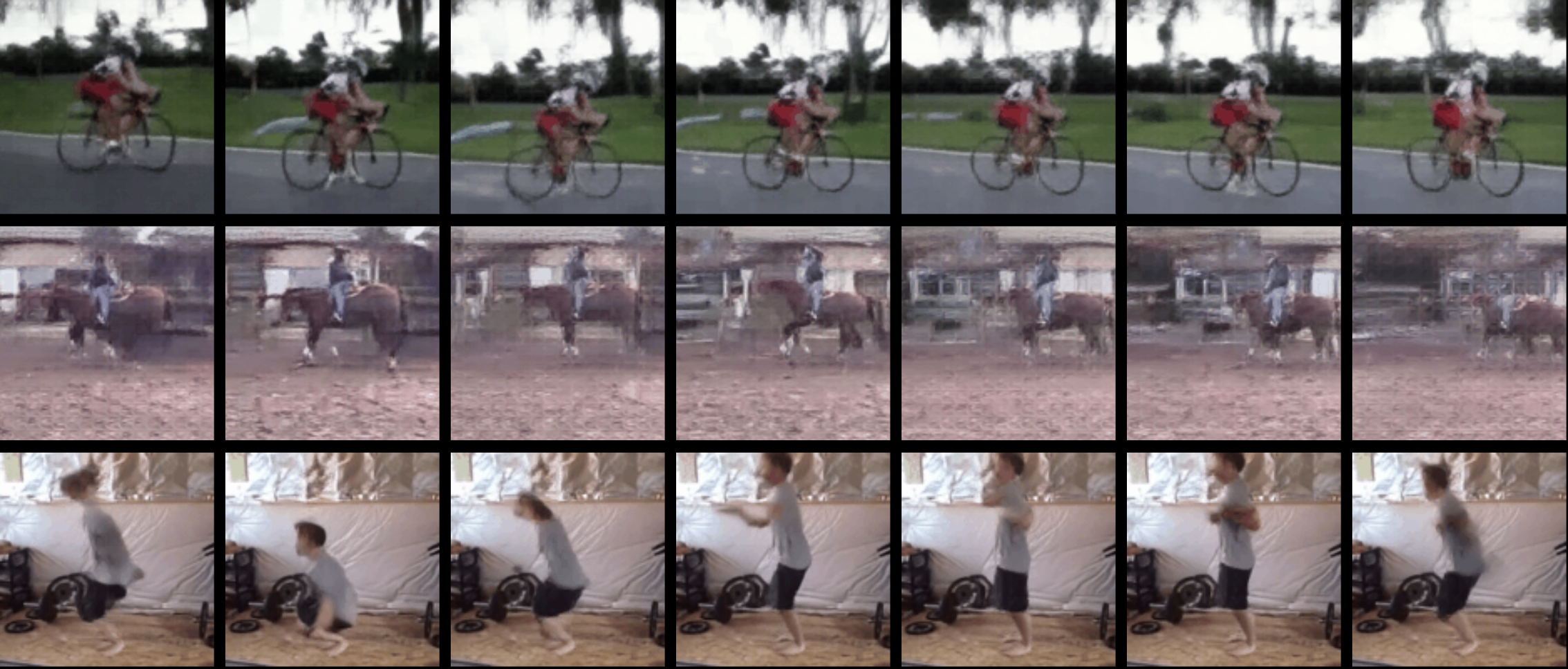}
    \caption{\textbf{512 Token Video Generation Examples.} We show randomly selected 16$\times$128$\times$128 video generation examples from our DiT-L trained at 512 tokens using the B-B/4x8 variant auto-encoder. Videos are sampled with 250 steps and a CFG weight of 2.0.}
    \label{fig:512_gen_results}
\end{figure}

\begin{figure}[h]
    \centering
    \includegraphics[width=0.95\textwidth]{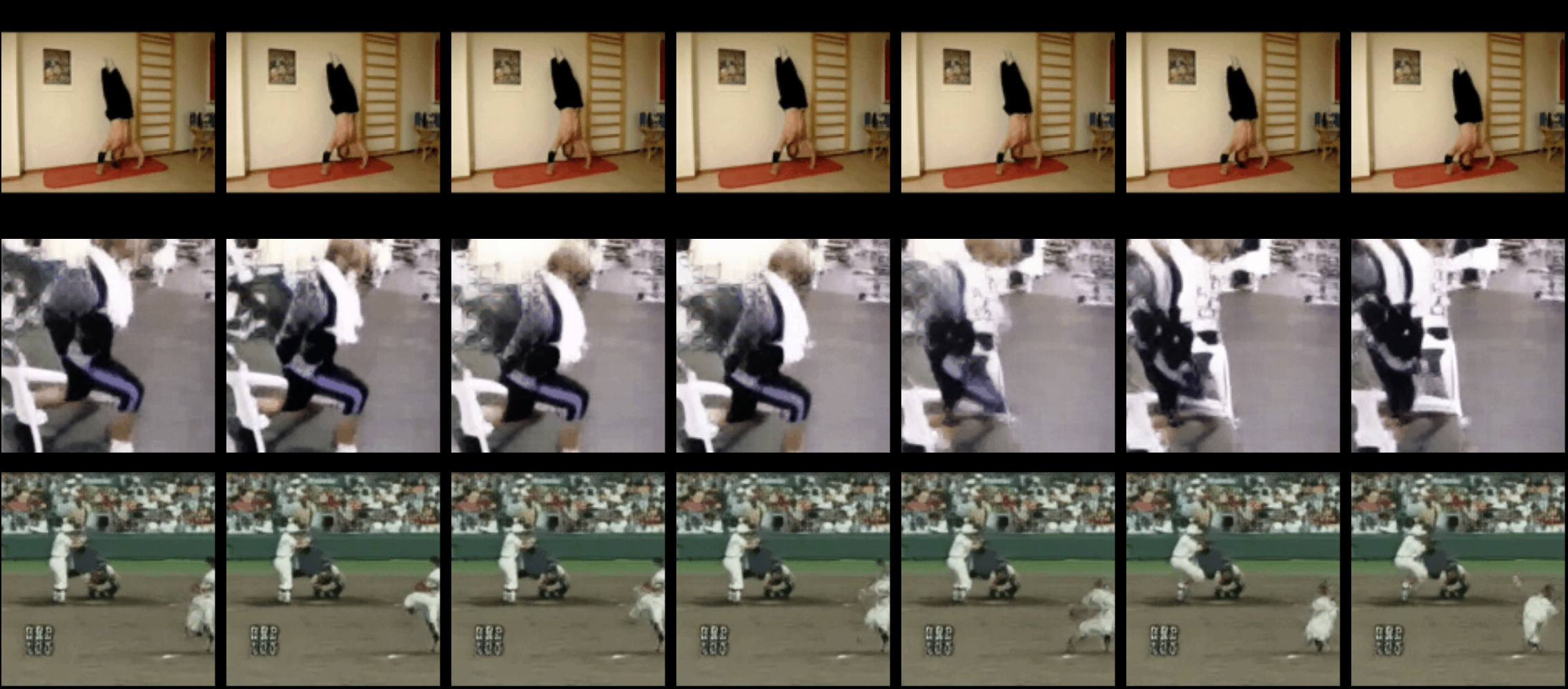}
    \caption{\textbf{256 Token Video Generation Examples.} We show randomly selected 16$\times$128$\times$128 video generation examples from our DiT-L trained at 256 tokens using the B-B/4x8 variant auto-encoder. Videos are sampled with 250 steps and a CFG weight of 2.0.}
    \label{fig:256_gen_results}
\end{figure}

\subsection{Image Sweep Generation Examples}
\begin{figure}[t]
\centering
\begin{tabular}{ccccccc}
\toprule
\multicolumn{7}{c}{Patch Size 8, Channel 4} \\
\multicolumn{7}{c}{\includegraphics[width=0.9\linewidth]{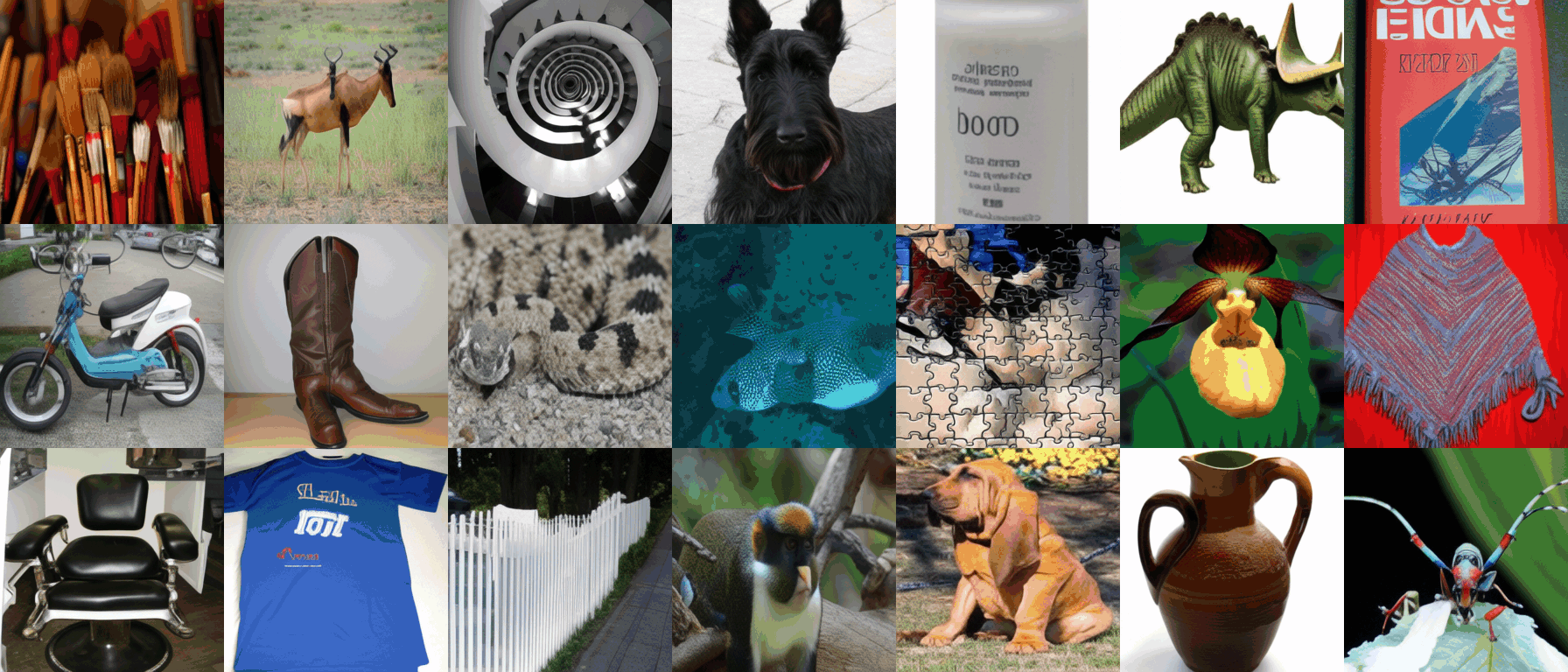}} \\
\midrule
\multicolumn{7}{c}{Patch Size 8, Channel 16} \\
\multicolumn{7}{c}{\includegraphics[width=0.9\linewidth]{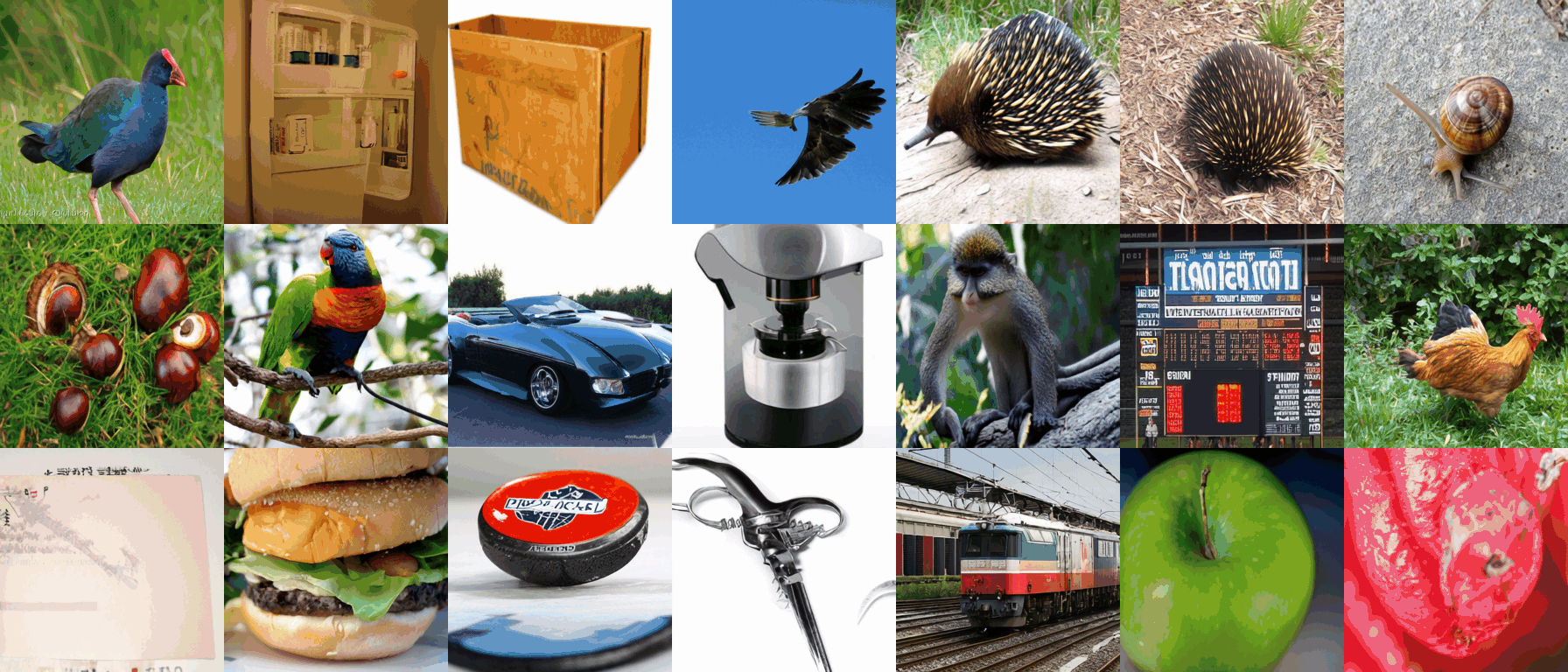}} \\
\midrule
\multicolumn{7}{c}{Patch Size 8, Channel 64} \\
\multicolumn{7}{c}{\includegraphics[width=0.9\linewidth]{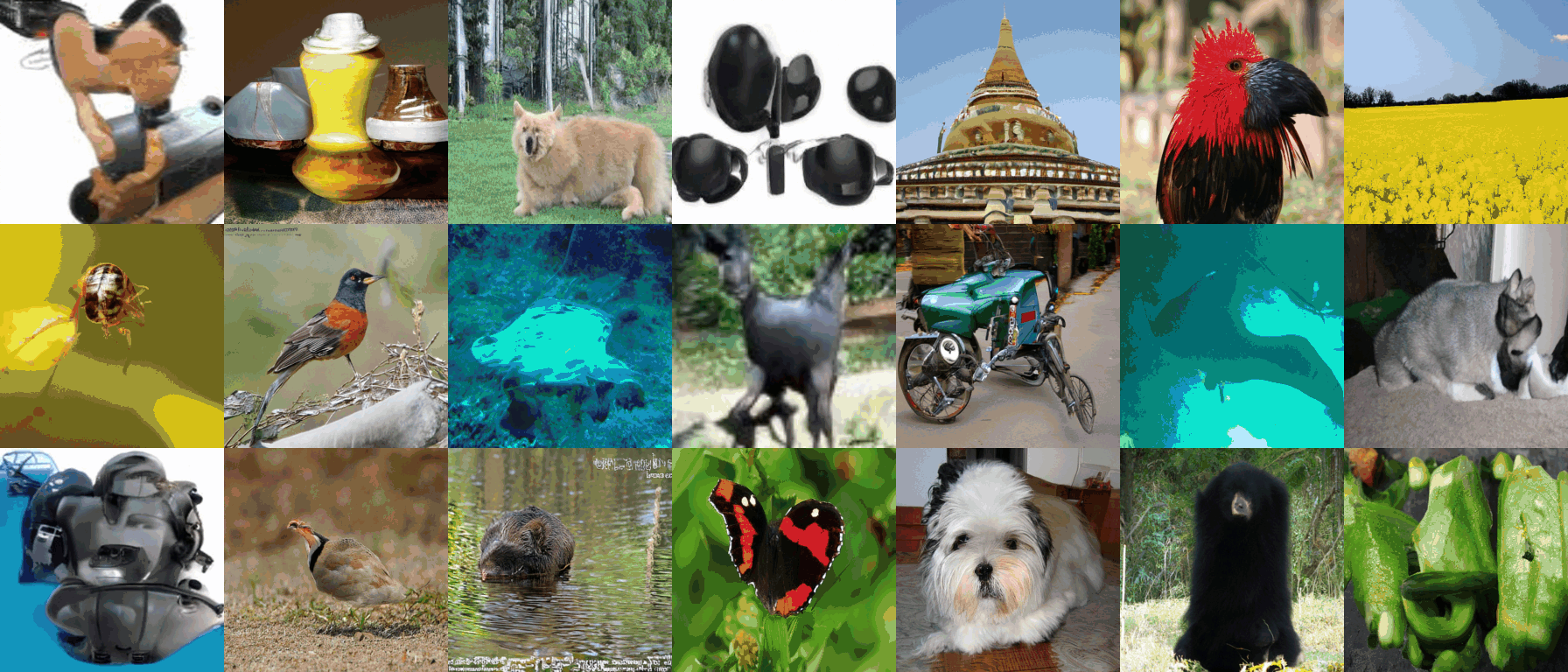}} \\
\bottomrule
\end{tabular}
\caption{\label{fig:gen_viz_1} \textbf{Channel size generation visualization 256p for $p=8$.} We show example generations for various compression ratios on \method S-B/8 from Figure~\ref{fig:256_gen_results}. Here $c=4$ has the best visuals that look close to good images, while $c=16$ generally looks good as well but not as good. $c=64$ looks very poor and the images do not look realistic.
}
\end{figure}

\begin{figure}[t]
\centering
\begin{tabular}{ccccccc}
\toprule
\multicolumn{7}{c}{Patch Size 16, Channel 4} \\
\multicolumn{7}{c}{\includegraphics[width=0.9\linewidth]{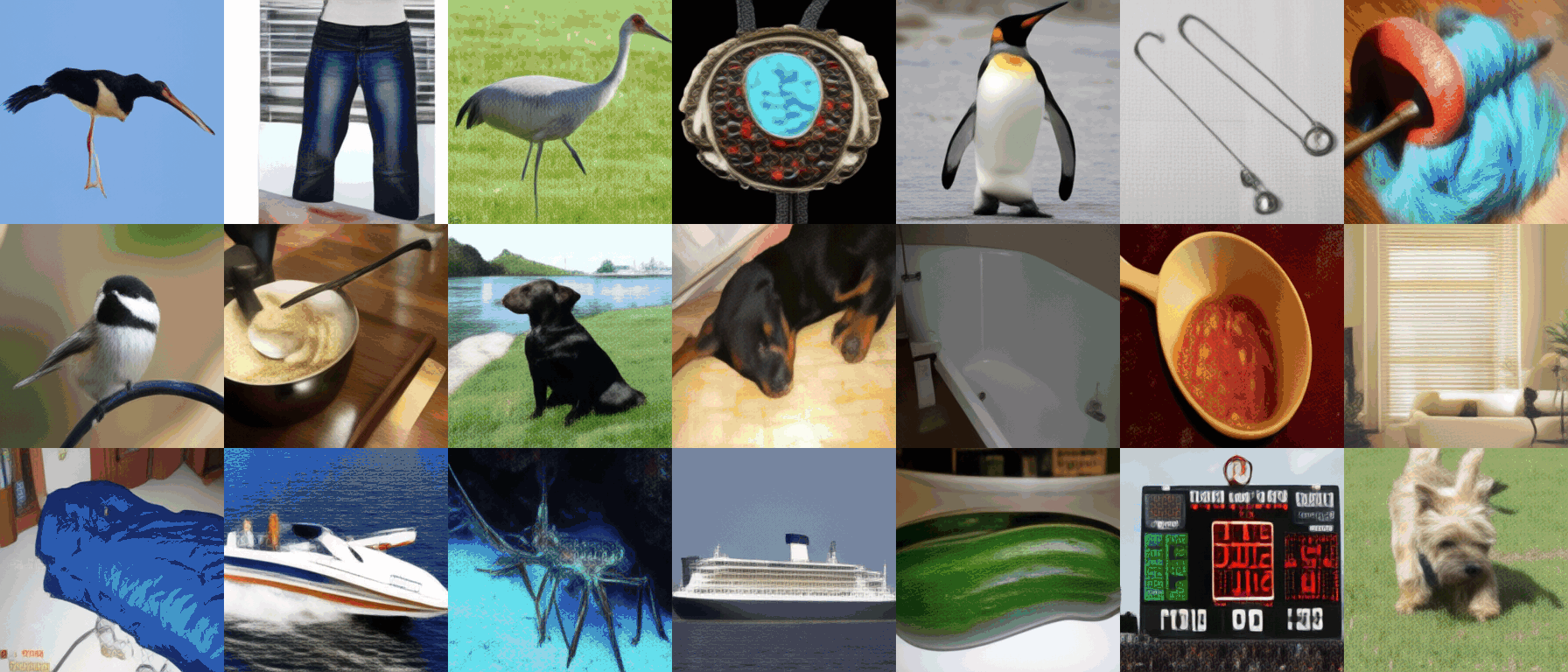}} \\
\midrule
\multicolumn{7}{c}{Patch Size 16, Channel 16} \\
\multicolumn{7}{c}{\includegraphics[width=0.9\linewidth]{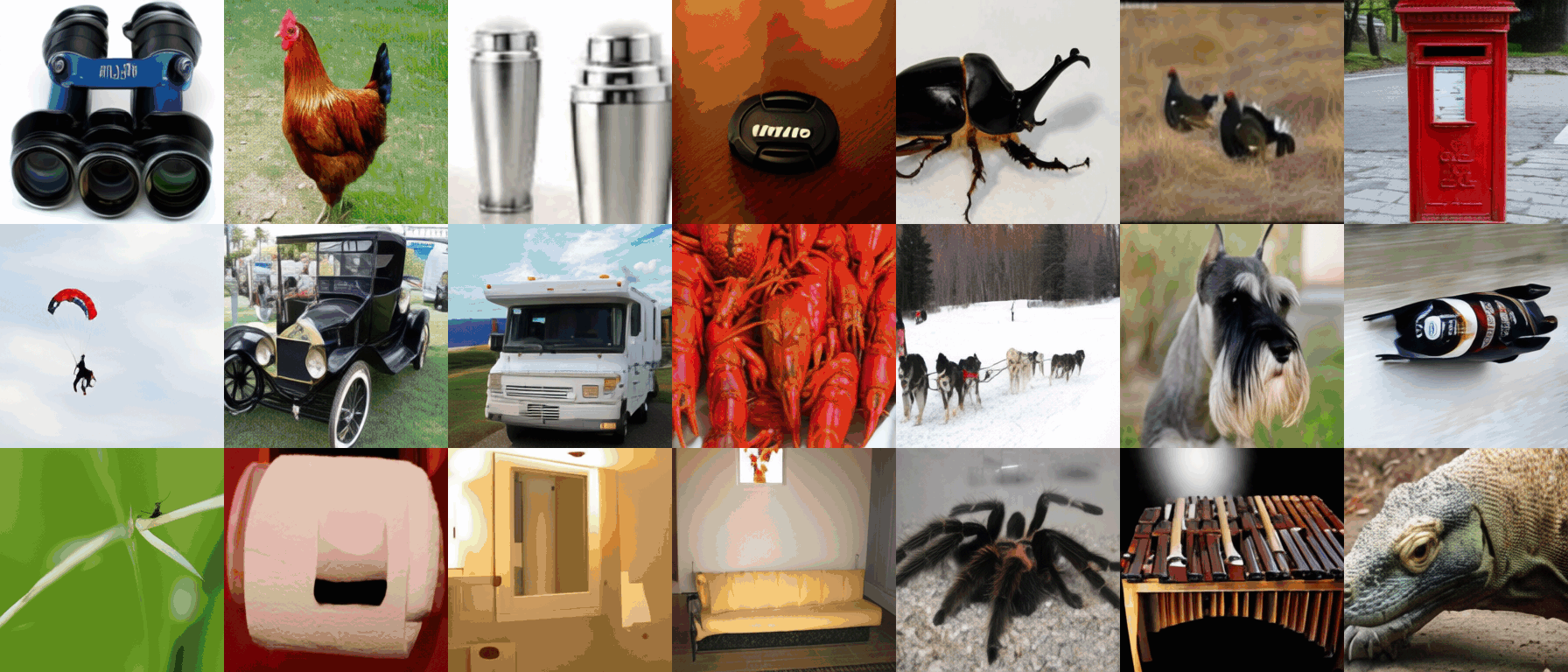}} \\
\midrule
\multicolumn{7}{c}{Patch Size 16, Channel 64} \\
\multicolumn{7}{c}{\includegraphics[width=0.9\linewidth]{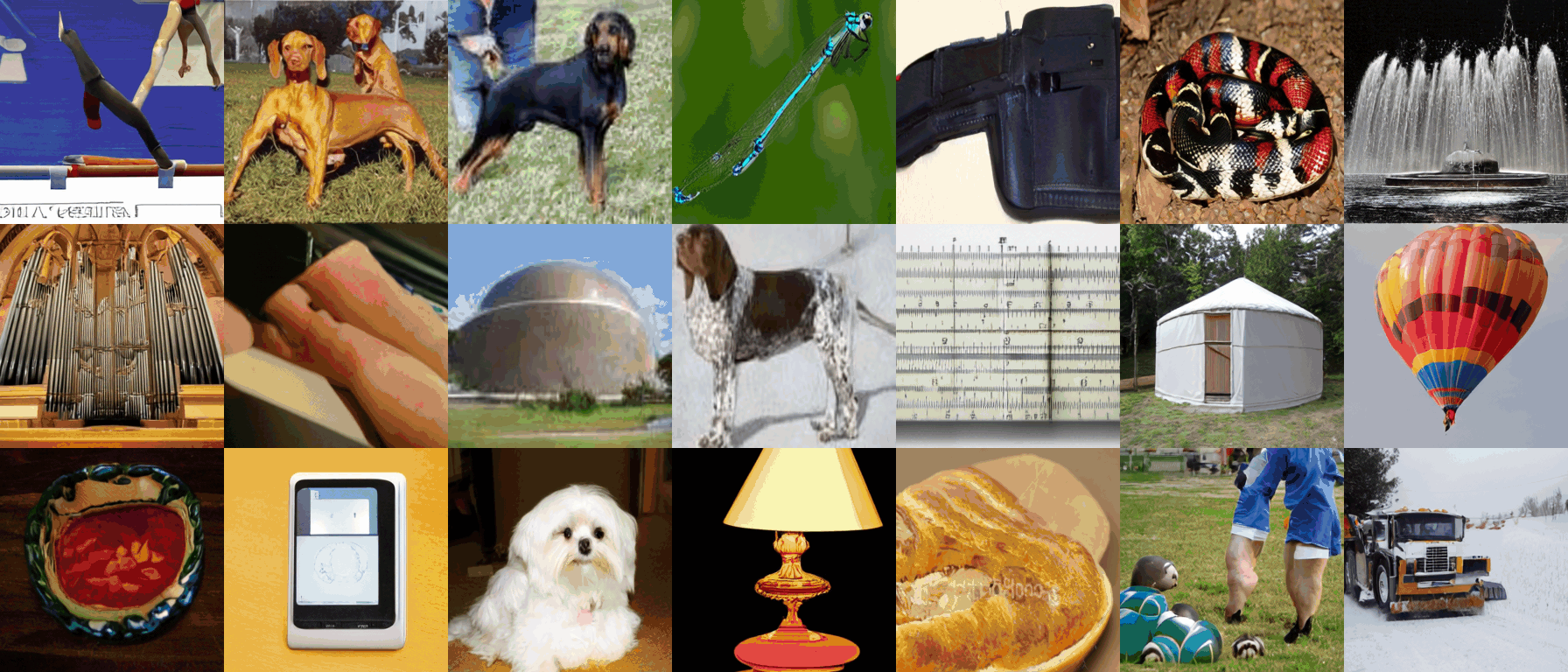}} \\
\bottomrule
\end{tabular}
\caption{\label{fig:gen_viz_2} \textbf{Channel size generation visualization 256p for $p=16$.} We show example generations for various compression ratios on \method S-B/16 from Figure~\ref{fig:256_gen_results}. Here $c=16$ has the best visuals that look close to good images, while $c=64$ suffers artifacts that worsen image quality. $c=4$ suffers from poor reconstruction quality from the auto-encoder.
}
\end{figure}

\begin{figure}[t]
\centering
\begin{tabular}{ccccccc}
\toprule
\multicolumn{7}{c}{Patch Size 32, Channel 4} \\
\multicolumn{7}{c}{\includegraphics[width=0.9\linewidth]{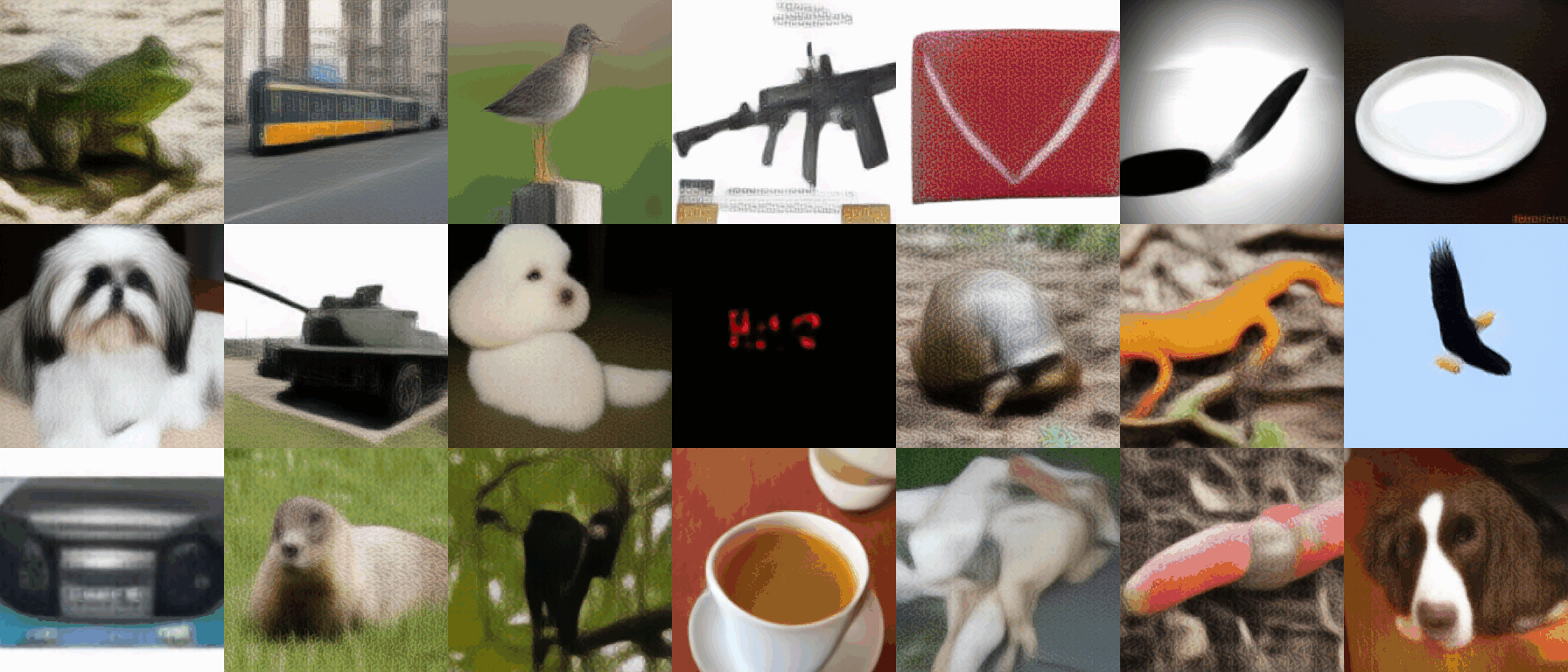}} \\
\midrule
\multicolumn{7}{c}{Patch Size 32, Channel 16} \\
\multicolumn{7}{c}{\includegraphics[width=0.9\linewidth]{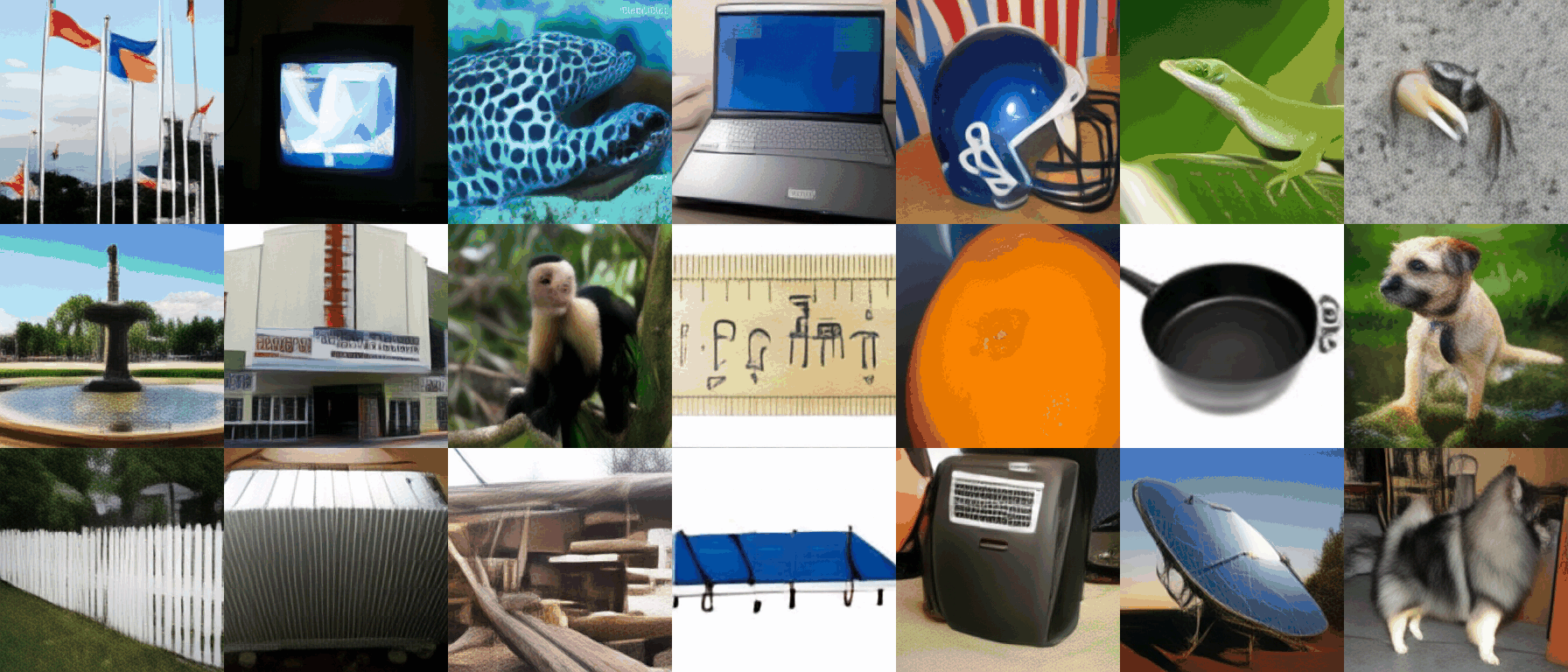}} \\
\midrule
\multicolumn{7}{c}{Patch Size 32, Channel 64} \\
\multicolumn{7}{c}{\includegraphics[width=0.9\linewidth]{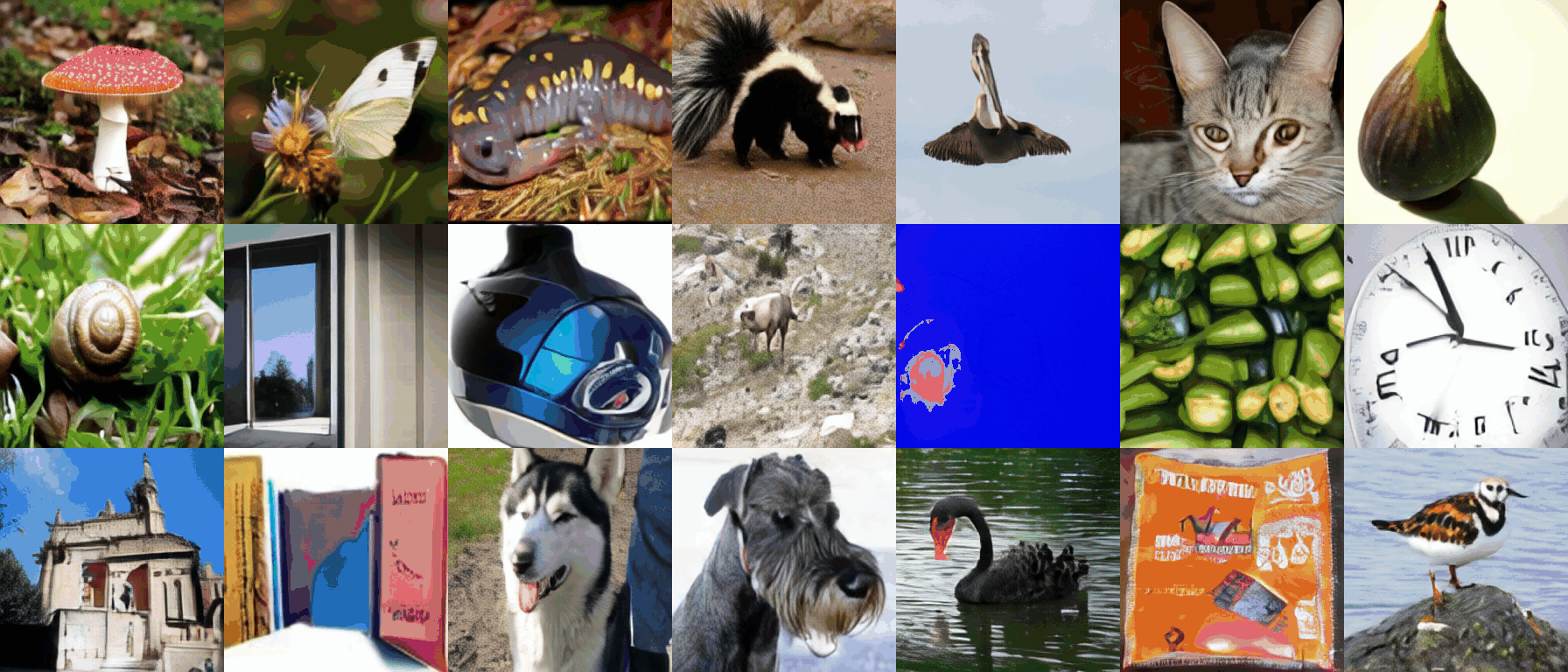}} \\
\bottomrule
\end{tabular}
\caption{\label{fig:gen_viz_3} \textbf{Channel size generation visualization 256p for $p=32$.} We show example generations for various compression ratios on \method S-B/32 from Figure~\ref{fig:256_gen_results}. Here $c=64$ has the best visuals overall but the high channel sizes make the image quality look poor and jumbled. Both $c-16$ and $c=4$ suffers from poor reconstruction quality from the auto-encoder.
}
\end{figure}

Here provide generation examples from our sweep conducted in Figure~\ref{fig:256_gen_results}. $p=16$ visuals are in Figure~\ref{fig:gen_viz_1}, $p=32$ visuals are in Figure~\ref{fig:gen_viz_2}, and $p=8$ visuals are in Figure~\ref{fig:gen_viz_3}.

\end{document}